\documentclass[letterpaper, journal]{IEEEtran}

\usepackage{hyperref}
\usepackage{cite}

\usepackage{amsmath,amssymb,amsfonts}
\usepackage{multirow}

\usepackage{graphicx}
\usepackage{tabularx}
\usepackage{subcaption} 
\usepackage{textcomp}
\usepackage{siunitx}
\usepackage{xcolor}
\usepackage{balance}
\usepackage{bm}
\usepackage{algorithmic}
\usepackage[linesnumbered,ruled,vlined]{algorithm2e}
\usepackage{algorithm2e}
\usepackage{subcaption} 
\newtheorem{mydef}{\textbf{Definition}}
\newtheorem{myremark}{\textbf{Remark}}

\newtheorem{myassumption}{\textbf{Assumption}}

\title{\LARGE \bf
Safe Dynamic Motion Generation in Configuration Space Using Differentiable Distance Fields
}

\author{Xuemin Chi$^{1,2}$, Yiming Li$^2$, Jihao Huang$^{1}$, Bolun Dai$^3$, Zhitao Liu$^{1\dagger}$, Sylvain Calinon$^{2\dagger}$%
\thanks{This work was supported in part by National Key R\&D Program of China (Grant NO. 2021YFB3301000); National Natural Science Foundation of China (NSFC:62173297), Zhejiang Key R\&D Program (Grant NO. 2022C01035).}%
\thanks{$^1$ Author is with the State Key Laboratory of Industrial Control Technology, Institute of Cyber-Systems and Control, Zhejiang University, Hangzhou, China. \tt\footnotesize \{Xuemin Chi, jihaoh, ztliu\}@zju.edu.cn}
\thanks{$^2$ Author is with Idiap Research Institute, 1920 Martigny, Switzerland, and also with Ecole Polytechnique Fédérale de Lausanne (EPFL), 1015 Lausanne, Switzerland. \tt\footnotesize \{yiming.li, sylvain.calinon\}@idiap.ch}
\thanks{$^3$ Author is with Fauna Robotics, New York, New York, United States of America. \tt\footnotesize bolundai16@gmail.com}
\thanks{$^\dagger$ Corresponding author.} 
\thanks{This letter has supplementary material available at \url{https://youtu.be/qscs9E36lrs}.}
}

\newcommand{\m}{\, \si[per-mode=symbol]{\metre}}
\newcommand{\rad}{\, \si[per-mode=symbol]{\radian}}
\newcommand{\mps}{\, \si[per-mode=symbol]{\metre\per\second}}

\begin{document}
\maketitle
\thispagestyle{empty}
\pagestyle{empty}

\begin{abstract}
Generating collision-free motions in dynamic environments is a challenging problem for high-dimensional robotics, particularly under real-time constraints.
Control Barrier Functions (CBFs), widely utilized in safety-critical control, have shown significant potential for motion generation. 
However, for high-dimensional robot manipulators, existing QP formulations and CBF-based methods rely on positional information, overlooking higher-order derivatives such as velocities.
This limitation may lead to reduced success rates, decreased performance, and inadequate safety constraints.
To address this, we construct time-varying CBFs (TVCBFs) that consider velocity conditions for obstacles.
Our approach leverages recent developments on distance fields for articulated manipulators, a differentiable representation that enables the mapping of objects' position and velocity into the robot's joint space, offering a comprehensive understanding of the system's interactions.
This allows the manipulator to be treated as a point-mass system thus simplifying motion generation tasks.
Additionally, we introduce a time-varying control Lyapunov function (TVCLF) to enable whole-body contact motions.
Our approach integrates the TVCBF, TVCLF, and manipulator physical constraints within a unified QP framework.
We validate our method through simulations and comparisons with state-of-the-art approaches, demonstrating its effectiveness on a 7-axis Franka robot in real-world experiments.
\end{abstract}

\section{Introduction}
Robots are expected to respond quickly and safely in dynamic environments.
For high-dimensional systems, generating safe motions at high frequencies is particularly challenging~\cite{koptev2024reactive}.
A hierarchical approach is predominantly applied in static environment scenarios, where a globally feasible path is planned offline, followed by full-horizon trajectory optimization to smooth the path while accounting for more complex constraints.
The trajectory optimization process is commonly formulated as a nonlinear programming problem (NLP) and is solved using gradient-based solvers.
Techniques such as CHOMP~\cite{ratliff2009chomp} and TrajOpt~\cite{schulman2014motion}  are effective but may not respond quickly enough to the demands of dynamic environments which require faster reactions.

Model predictive control (MPC) methods can address this issue by using shorter horizons.
The formulation is often an NLP problem that requires an initial guess.
This includes traditional gradient-based MPC~\cite{spahn2021coupled} and recent gradient-free MPC using sampling-based methods, e.g. MPPI~\cite{williams2016aggressive} and VP-STO~\cite{jankowski2023vp}.
However, the replanning frequencies of these algorithms are struggling to balance safety and efficiency.
The computation of system dynamics and nonlinear constraints for each rollout state grows exponentially.
A longer horizon allows a higher chance of finding a feasible solution, as MPC can ensure safety within the preview window if a solution exists, but this comes at the cost of higher computational demands. 

The quadratic programming (QP) formulation provides real-time efficiency for high-dimensional systems and has become popular in recent years~\cite{koptev2021real,mirrazavi2018unified}.
Recent works have explored the formulation of QP controllers for motion generation tasks~\cite{koptev2024reactive,li2024configuration}, by modeling the geometry of the robot as a signed distance field (SDF) ~\cite{koptev2022neural, li2024representing}, achieve reactive motion policies
with a high frequency of over 200 Hz.
These methods typically consider the positional relationship of objects and the robot, ignoring high-order information such as velocities.
Such limitation prevents the robot from reacting appropriately to objects. 
For instance, the robot's motion should differ based on the object's speed and direction, particularly when the object is approaching or away from the robot.
As a result, the robot's behavior usually relies on manually set parameters, limiting its applicability in dynamic scenarios and the safety behavior is not guaranteed.

To address this challenge, it is crucial to describe the object's dynamics, including both position and velocity information.
In this paper, we propose to exploit the paradigm of control barrier functions to address this challenge, allowing the robot to react safely based on the objects' dynamic states.
Such an approach has been explored in task space-based planning fields such as autonomous driving~\cite{zeng2021safety}, providing a simple and efficient approach for safety control synthesis.
However, this task becomes more complex when applied to high-dimensional robotic manipulators due to the nonlinear mapping of geometries in task and joint spaces.
Thus, previous studies typically construct CBFs based on the distance information~\cite{singletary2022safety,dai2023safe}. 

We propose to construct time-varying CBFs that incorporate the velocities information of objects.
This is achieved by leveraging recent advancements in distance fields for robot manipulators, a differentiable representation that can reason the geometry of obstacles in robot configuration space~\cite{li2024representing,li2024configuration}.
The derivative of the distance field corresponds to the velocity of obstacles in joint space.
This representation allows us to fully consider the articulated robot and the surrounding environment as a point-mass system, simplifying the motion generation problem.
We demonstrate a paradigm that synthesizes a safety-critical controller in configuration space.
Additionally, we design a time-varying CLF to address a whole-body dynamic reaching task.
The primary contributions of this work are as follows:
\begin{itemize} 
    \item We introduce an approach to map the position and velocity of the objects to robot configuration by leveraging the differentiable distance field representation of robot manipulators.

    \item We construct a singularity-free TVCLF to facilitate whole-body reaching tasks in configuration space.
    Additionally,
    TVCBFs are developed to ensure safety-aware motion generation.
    The constraints of TVCLFs, TVCBFs and physical limits are systematically formulated as a QP.
    
    \item We validate our approach through numerical simulations on planar robot arms, with benchmarks against baselines providing a detailed performance analysis.
    To further validate the proposed approach, we performed simulations and real-robot experiments using a 7-axis Franka robotic arm.
\end{itemize}

The rest of this paper is organized as follows:
In Section~\ref{sec:background}, we briefly review CBF, CLF, and the formulation of the CBF-CLF-QP.
Section~\ref{sec:problem formulation} defines the robotic control problem.
In Section~\ref{sec:method}, we detail the construction of the TVCBFs, TVCLF, and the integrated QP formulation.
Section~\ref{sec:experiments} presents simulations and real robot experiments.
Finally, in Section~\ref{sec:conclusion}, we conclude the paper with a discussion of results and directions for future work.
\section{Background}\label{sec:background}
Our safety-critical controller builds on top of the control Lyapunov functions and control barrier functions.
We present the necessary preliminaries in this section.

\subsection{Distance Fields for Robot Manipulators}
Signed distance fields (SDFs) are popular representations in robotics in modeling environments.
Recent work extends this concept to articulated robots, by either exploiting neural networks to approximate the distance function~\cite{koptev2022neural,liu2022regularized} or leveraging the kinematics chain of the robot~\cite{li2024representing}:

\begin{equation}
    \label{eq:def_signed_distance}
    d = f(p,q),
\end{equation}
where $d\in \mathbb{R}$ is the distance value, $p\in \mathbb{R}^3$ and $q\in \mathbb{R}^n$ are the spatial point and the joint angles respectively.
Such representations are differentiable, enabling efficient distance and gradient queries for arbitrary point and joint configuration pairs.
In~\cite{li2024configuration}, the authors introduce configuration space distance fields (CDFs), which also estimate the distance given a specific point and the robot's configuration.
However, instead of modeling the task space distance from a point to the robot's surface, CDF measures the distance that indicates the minimal joint motion required by the robot to establish contact with the point.
In other words, this representation measures the distance in joint space by implicitly solving a whole-body inverse kinematics task.

\subsection{Control Lyapunov Functions Based QP}
Control Lyapunov functions are proposed to stabilize a system with a feedback control law.
Consider a nonlinear control-affine system
\begin{equation}
    \label{eq:def_control_affine_system}
    \dot{x} = f(x) + g(x)u,
\end{equation}
where the state $ x \in D \subset \mathbb{R}^n $ and the control input $ u \in U \subset \mathbb{R}^m $. Here, $ U $ denotes the admissible control set, defined as $ U = \{ u \in \mathbb{R}^m \mid u_{\min} \leq u \leq u_{\max} \} $, and $ D $ denotes the admissible state set, given by $ D = \{ x \in \mathbb{R}^n \mid x_{\min} \leq x \leq x_{\max} \} $.
The drift term $ f: \mathbb{R}^n \rightarrow \mathbb{R}^n $ and the control influence matrix $ g: \mathbb{R}^n \rightarrow \mathbb{R}^{n \times m} $ are both locally Lipschitz continuous.
\begin{mydef}[Classes $\mathcal{K}$ and $\mathcal{K}_{\infty}$ Functions~\cite{ames2019control}]
    A Lipschitz continuous function $\mu: [0, a) \to [0, \infty)$, where $a > 0$, is classified as a $\mathcal{K}$ function if it is strictly increasing and satisfies $\mu(0) = 0$.
    Additionally, a function is in class $\mathcal{K}_{\infty}$ if it meets the criteria of class $\mathcal{K}$, with the further properties that $a = \infty$ and $\mu(b) \to \infty$ as $b \to \infty$.
    \label{def:class_function}
\end{mydef}

A continuously differentiable function $V$ is a CLF for the system~\eqref{eq:def_control_affine_system} if it is positive definite and satisfies~\cite{ames2019control}
\begin{equation}
    \inf _{u \in U}\left[L_f V(x)+L_g V(x) u\right] \leq-\gamma(V(x)),
    \label{eq:define_clf}
\end{equation}
where $L_f V(x) := \frac{\partial V}{\partial x} f(x)$ and $L_g V(x) := \frac{\partial V}{\partial x}g(x)$ are Lie-derivatives of $V(x)$, $\gamma(\cdot)$ belongs to class $\mathcal{K}$.

Since CLF constraints are affine in controls, we can define a QP and the objective function is minimizing the control efforts:
\begin{align}
    u^{*}(x) = &\underset{u\in U}{\operatorname{argmin}} 
    ~\frac{1}{2} u^{\top} R u \\
    &\text{s.t.}~ L_f V(x)+L_g V(x) u \leq -\gamma(V(x)) \tag{CLF}
\end{align}
\label{eq:clf_qp_problem}
This QP-based controller is denoted as CLF-QP.

\subsection{Control Barrier Functions  Based QP}
\begin{mydef}[Forward Invariance]
    Let \( x(t) \) denote the unique solution to~\eqref{eq:def_control_affine_system} starting from an initial state \( x_0 \in \mathcal{C} \).
    The controller \( \pi(x) \) renders the system~\eqref{eq:def_control_affine_system} safe with respect to \( \mathcal{C} \) if it ensures that \( x(t) \) remains within the safe set \( \mathcal{C} \) for all \( t \in I(x_0) \), where \( I(x_0) = [t_0, t_{\max}) \) denotes the maximum interval of existence of \( x(t) \).
    \label{def:safe_set}
\end{mydef}

The safety constraints in~\eqref{eq:def_control_affine_system} can be framed as enforcing the forward invariance of a set, i.e., the system stays inside the safe set.
Consider a set $\mathcal{C}$ defined as the superlevel set of a continuously differentiable function $h: D \subset \mathbb{R}^{n} \rightarrow \mathbb{R}$:
\begin{equation}
    \label{eq:safe_set}
    \mathcal{C} = \{x\in D \subset \mathbb{R}^{n} \mid h(x) \geq 0 \}
\end{equation}
Throughout this paper, we refer to $\mathcal{C}$ as a safe set.
Consider we have access to a Lipschitz continuous controller $ u = \pi (x)$ such that $\dot{x}:= f(x) + g(x)\pi (x)$.

A continuously differentiable function \( h \) is called a CBF if it satisfies \( \frac{\partial h}{\partial x} \neq 0 \) for all \( x \in \partial \mathcal{C} \) and if, for the system~\eqref{eq:def_control_affine_system}, the following condition holds:
\begin{equation}
    \sup_{u \in U} \left[ L_f h(x) + L_g h(x) u \right] \geq -\alpha(h(x)),
    \label{eq:define_cbf}
\end{equation}
where \( L_f h(x) = \frac{\partial h(x)}{\partial x} f(x) \) and \( L_g h(x) = \frac{\partial h(x)}{\partial x} g(x) \) are the Lie derivatives of \( h(x) \) with respect to \( f \) and \( g \), respectively. Here, \( \alpha(\cdot) \) is an extended class \( \mathcal{K} \) function.

The CBF constraint, being affine in the control input \( u \), can be integrated directly into a QP alongside the CLF constraint:
\begin{align}
\label{eq:cbf_clf_qp_problem}
    u^{*}(x) = &\underset{u, \delta }{\operatorname{argmin}} 
    ~\frac{1}{2} u^{\top} R u + p \delta^2\\
    \text{s.t.}~ & L_f V(x) + L_g V(x) u + \gamma(V(x)) \leq \delta  \tag{CLF} \\
    & L_{f} h(x) + L_{g} h(x) u \geq -\alpha(h(x)) \tag{CBF} \\ 
    & u \in U,~\delta \in \mathbb{R_{+}}, \nonumber
\end{align}
where \( \delta \) is a relaxation variable introduced to prioritize safety over strict stability requirements, allowing the optimization problem to remain feasible for an appropriately chosen weight \( p > 0 \).
By penalizing \( \delta \) in the objective function, the system will only deviate from stabilization goals when necessary to satisfy safety constraints, maintaining a balance between the stability (CLF) and safety (CBF) objectives.

\section{Problem Formulation}\label{sec:problem formulation}
We aim to control an $n$-DOF manipulator with states being described by joint angles $q=[q_1, \ldots, q_n]$ to perform whole-body reaching and collision avoidance tasks in dynamic environments.
The obstacles to avoid and the object to reach are defined within the task space.
All joint angles are bounded by joint-limits $q\in [q_{\min}, q_{\max}]$.
The objective of this work is to design a safety-critical controller $\pi$ in configuration space where the manipulator system is linear and the geometry is a point that accomplishes the above tasks.

\section{method}\label{sec:method}
In this section, we detail the development of a velocity-based controller designed to enable collision-free, whole-body reaching motions.
The controller employs a TVCLF and TVCBFs to ensure task completion and safety.
We also describe our approach that leverages the distance field representation of the robot to obtain the corresponding joint velocity given moving objects, allowing for the construction of time-varying constraints.
Finally, we integrate physical constraints into the framework, embedding these constraints within a unified QP formulation.

\subsection{Design CLFs in Configuration Space}

\begin{figure*}[ht!]
    \centering
    \begin{subfigure}[t]{0.23\linewidth}
        \centering
        \includegraphics[width=\linewidth]{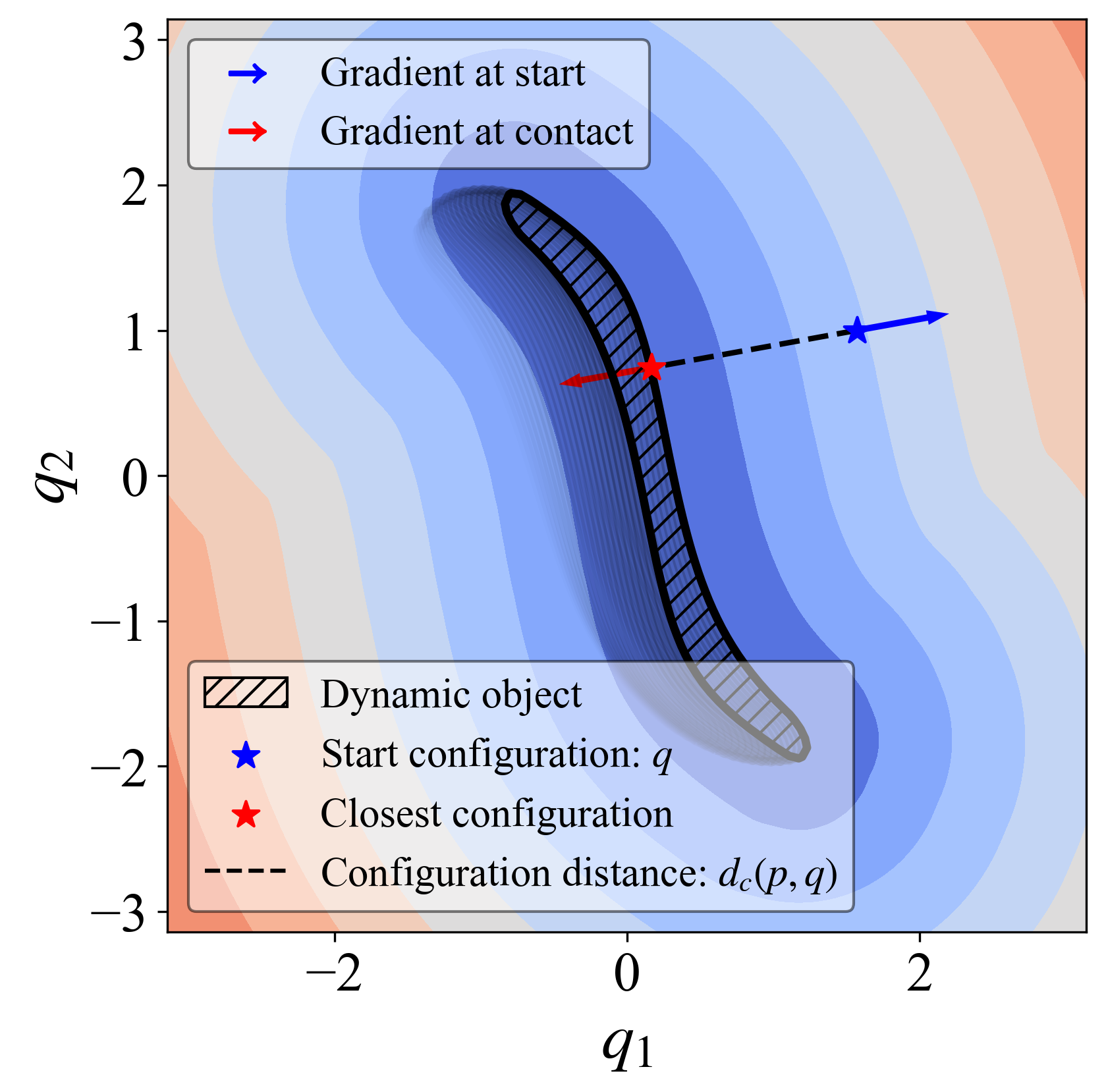}
        \caption{CDF: Configuration Space}
        \label{fig:cdf_c_space}
    \end{subfigure}%
    \hspace{0.02\linewidth} 
    \begin{subfigure}[t]{0.23\linewidth}
        \centering
        \includegraphics[width=\linewidth]{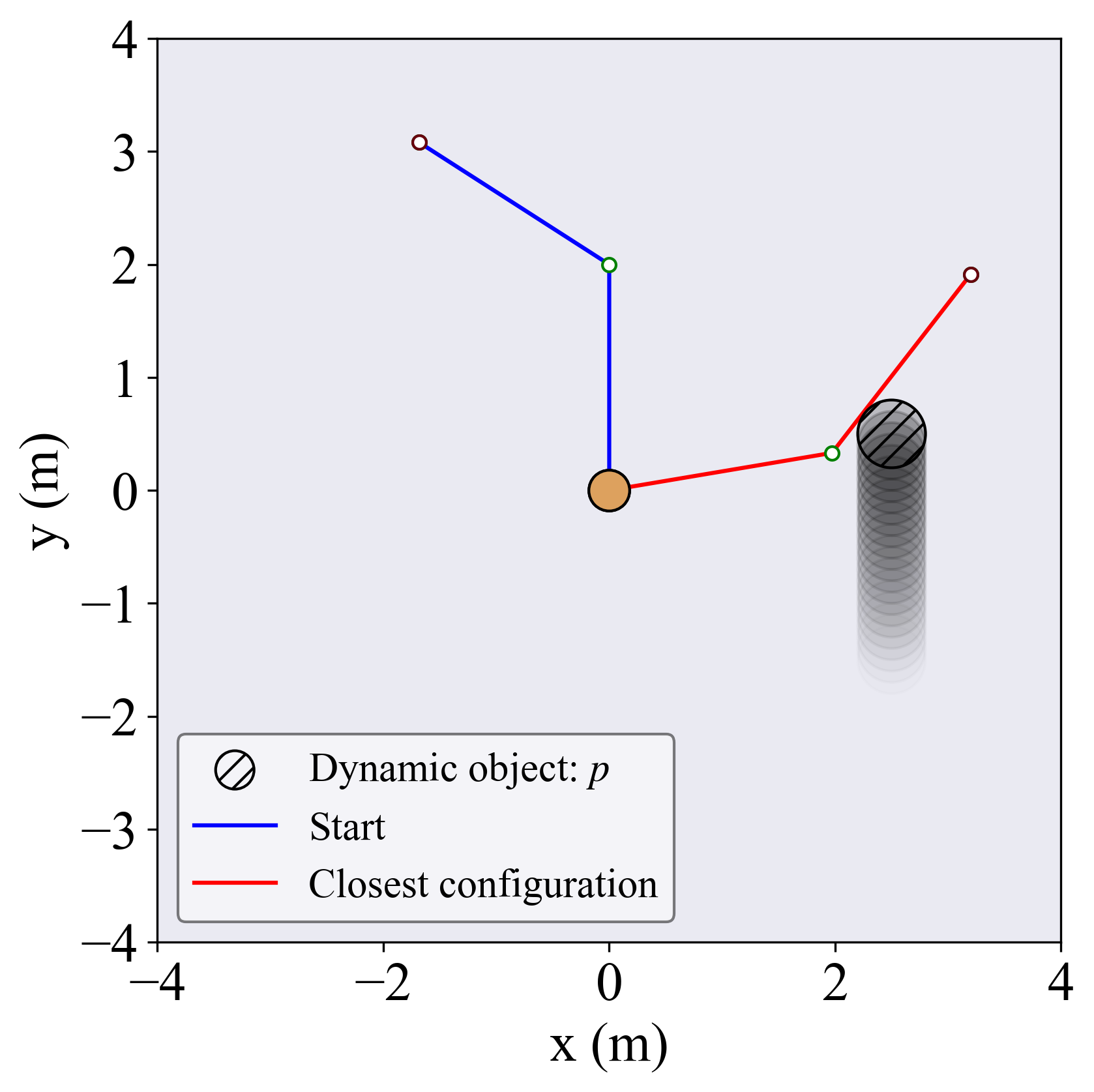}
        \caption{CDF: Task Space}
        \label{fig:cdf_t_space_reachable}
    \end{subfigure}%
    \hspace{0.02\linewidth}
    \begin{subfigure}[t]{0.23\linewidth}
        \centering
        \includegraphics[width=\linewidth]{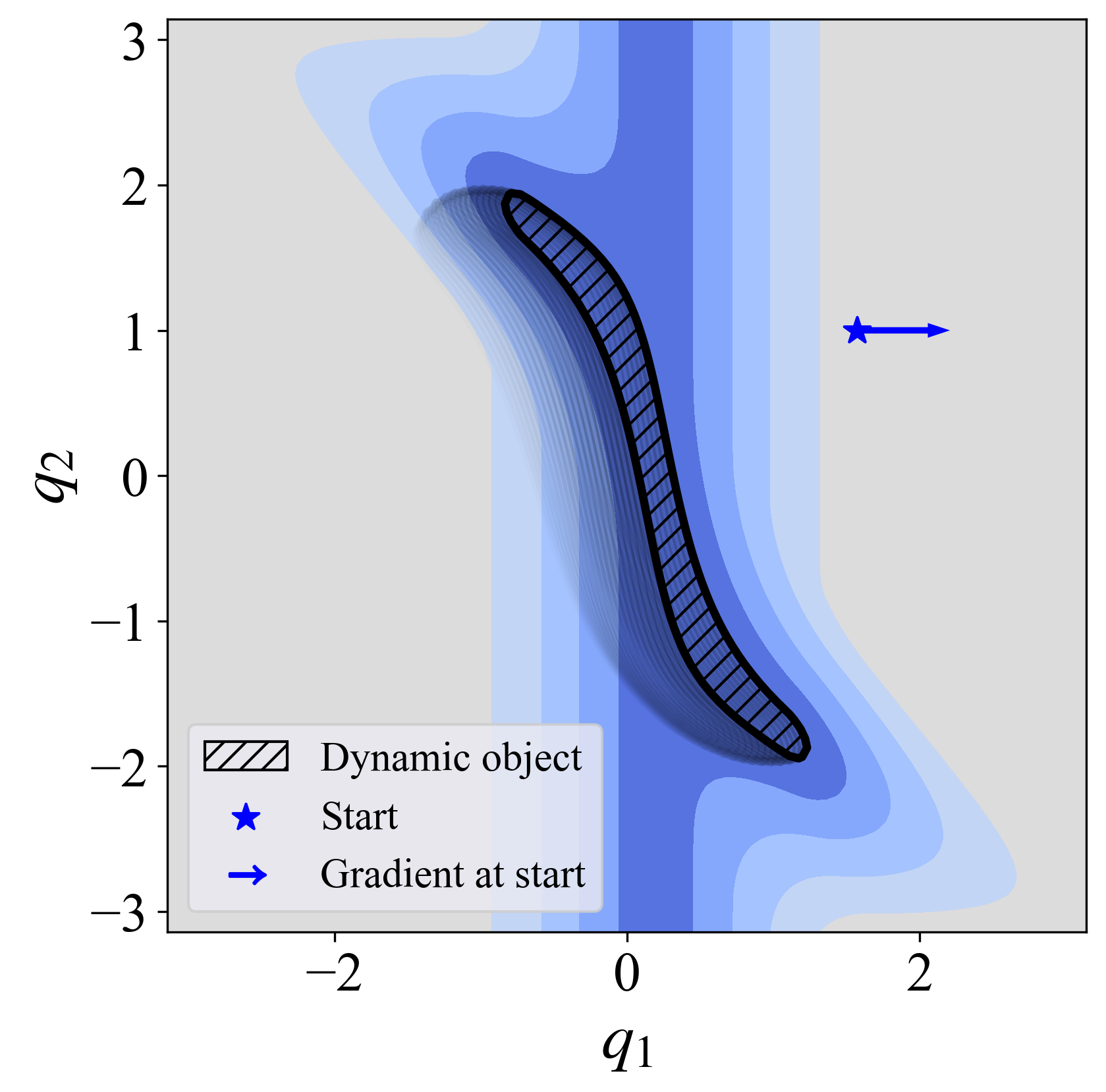}
        \caption{SDF: Configuration Space}
    \end{subfigure}%
    \hspace{0.02\linewidth}
    \begin{subfigure}[t]{0.23\linewidth}
        \centering
        \includegraphics[width=\linewidth]{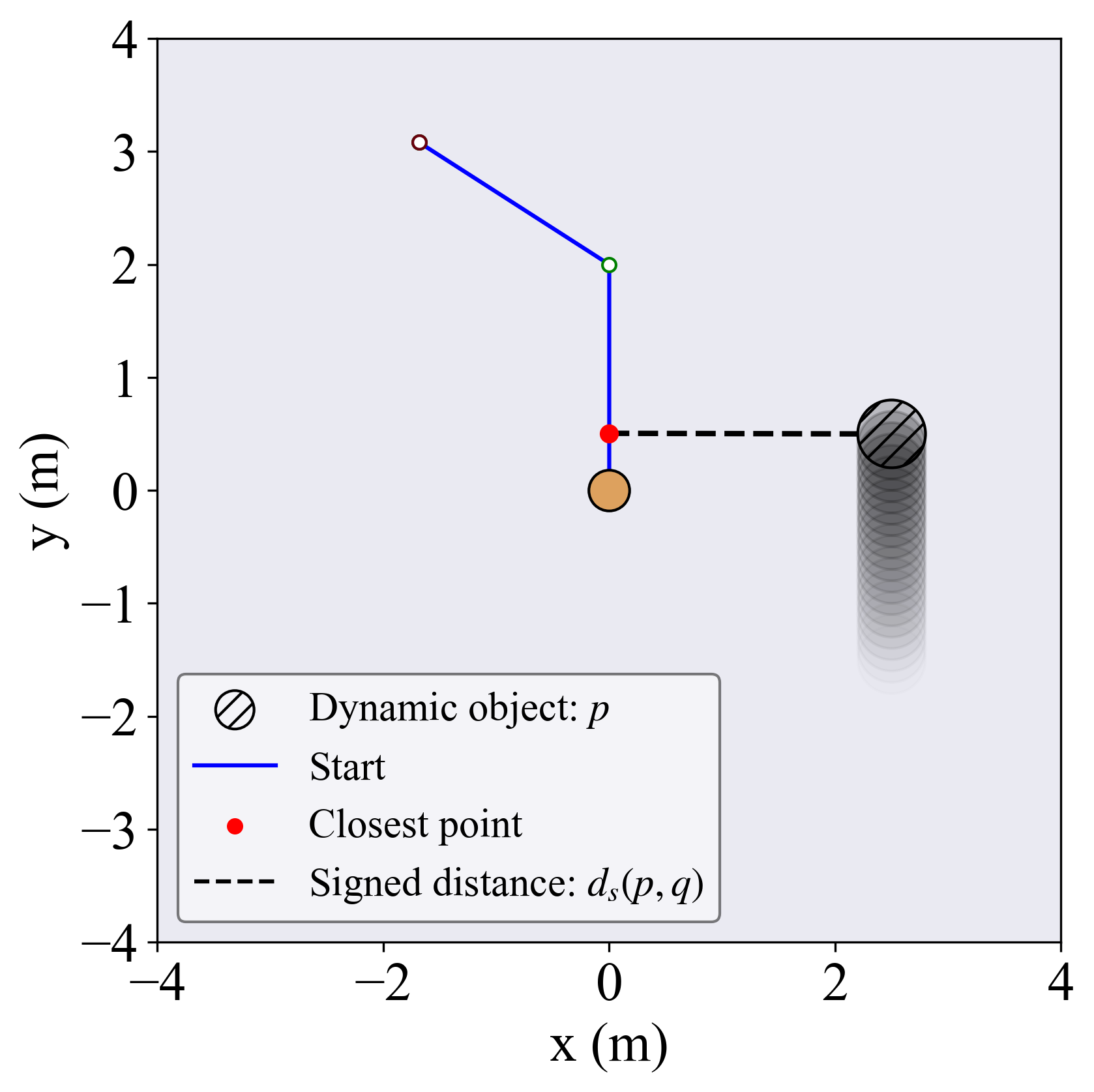}
        \caption{SDF: Task Space}
        \label{fig:sdf_t_space}
    \end{subfigure}
    \caption{The distance and gradient information of CDF and SDF.
    (a)~Object velocity captured in the configuration space.
    (b)~The object is reachable.
    (c)~Partial gradient information.
    (d)~The object velocity is directly employed in the task space.
    }
    \label{fig:sdf_cdf:demonstration}
\end{figure*}

We intend to design a TVCLF to stabilize our system at the target object.
The position of a moving object in task space is denoted as $ p_g (t) \in \mathbb{R}^{3}$.
The velocity of the moving object is predefined in task space and denoted as $v_g \in \mathbb{R}^{3}$.
To account for the object's motion, we define a TVCLF, which modifies the standard CLF condition~\eqref{eq:define_clf} as follows:
\begin{equation}
\label{eq:tv_clf}
\resizebox{\columnwidth}{!}{$
    \inf\limits_{u \in U} \left[L_f V(x, p_g(t)) + L_g V(x, p_g(t)) u + \frac{\partial V(x,p_g(t))}{\partial t}\right] \leq -\gamma(V(x, p_g(t)))
$}
\end{equation}
where $\frac{\partial V(x, p_g(t))}{\partial t}$ accounts for the object’s dynamics.

Following the notation in~\eqref{eq:def_signed_distance}, let $d_{c}(p_g, q)$ denote the configuration distance between the robot at joint configuration $q$ and the object at state $p_g$.
To ensure $d_{c}(p_g, q)$ meaningful, we make the following assumption:
\begin{myassumption}
    \label{assum:reachable}
    The object in the task space is considered reachable if the states of the manipulator are subjected to physical limits and the control commands are admissible as shown in Fig.~\ref{fig:cdf_t_space_reachable}.
\end{myassumption}

When the object's position $p_g$ is out of the task space, the concept of configuration distance is undefined and the configuration space for the object is empty.

The first term $L_f V(x, p_g(t)) = \frac{\partial V(x,p_g(t))}{\partial x}  f(x) \equiv \frac{\partial d_c({p_g, q})}{\partial q}  f(q)$ is zero since the system is modeled as a single integrator system, $\dot{q}=u$.

The second term in~\eqref{eq:tv_clf} follows $L_g V(x, p_g(t)) = \frac{\partial V(x,p_g(t))}{\partial x} g(x) \equiv \frac{\partial d_c\left(p_g, q\right)}{\partial q}  g(x)$, where $g(x)$ is an identity matrix, consistent throughout this paper.

In this paper, we propose a solution for the third term, $ \frac{\partial V(x, p_g(t))}{\partial t} $, addressing the challenge of bridging the task space and configuration space of the manipulator.
Specifically:
$
\frac{\partial V(x, p_g(t))}{\partial t} = \frac{\partial V(x, p_g(t))}{\partial p_g(t)} \frac{\partial p_g(t)}{\partial t} \equiv \frac{\partial d_{c}(p_g, q)}{\partial p_g} v_g.
$
Here, $\partial p_g$ measures position changes of the object in the task space, while  $\partial d_c(p_g, q)$ measures configuration space distance changes using CDF as shown in Fig.~\ref{fig:cdf_c_space}.
The term $\frac{\partial p_g}{\partial t}$ represents the velocity of the target object in the task space $v_g$.
The constructed time-varying constraint using $d_c(p_g, q)$, referred to as CDF-TVCLF, is summarized as follows:
\begin{equation}
\text{CDF-TVCLF}\left\{
\begin{aligned}
  L_f V(x, p_g(t)) &= 0 \\
  L_g V(x, p_g(t))&= \frac{\partial d_c\left(p_g, q\right)}{\partial q} g(x) \\
  \frac{\partial V(x, p_g(t))}{\partial t} &= \frac{\partial d_{c}(p_g, q)}{\partial p_g}  v_g.
  \label{eq:tvclf_formulation}
\end{aligned}
\right.
\end{equation}
$d_c(p_g, q)$ is a continuously differentiable, nonnegative function, and its derivative with respect to the joint configuration $q$ always has a unit norm. Therefore, the Lie-derivative $L_g V(x, p_g(t)$ is non-zero.

\begin{myremark}
    If the TVCLF constraint is enforced strictly (without relaxation), the QP may become infeasible when the constraint cannot be met. This situation implies that none of the manipulator’s links are able to reach the target. To address this, we allow the CLF constraint to be relaxed, which helps the QP remain feasible even if one part of the manipulator is unable to reach the target. In such cases, another link of the manipulator might still be able to achieve the goal.
\end{myremark}

By leveraging the CDF representation which leads to a unit gradient norm in joint space pointing to the target object, the proposed TVCLF overcomes singularity issues caused by forward kinematics.
Specifically, when the robot arm is fully stretched out and the target object is collinear with the manipulator, traditional methods—such as those relying on signed distances or task-space-based formulations—encounter zero-gradient issues, leading to singularities in whole-body reaching tasks or failures in end-effector reaching.

In the specific case of the collinear configuration, where the robot and target object are aligned, singularity occurs in traditional methods due to loss of directional information. However, by the definition of the CDF, a target pose $\hat{q}$ always exists such that  $d_c(p_g, \hat{q}) = 0$, minimizing the configuration distance.
Since the gradient is non-zero, the TVCLF designed using the CDF ensures that the manipulator is guided toward the target pose $\hat{q}$, overcoming the singularity issue.
Therefore, the method guarantees reaching the collinear target object as long as it is reachable in the task space.

\subsection{Design CBFs in Configuration Space}
This section addresses the design of time-varying CBFs (TVCBF) for velocity-controlled robotic systems operating in dynamic environments with moving obstacles.
Let there be $N_o$ moving obstacles in the task space, where each obstacle’s state is given by $p_i(t) \in \mathbb{R}^3$ for $i = 1, \dots, N_o$.
Since the obstacles are moving, the safe set~\eqref{eq:safe_set} becomes time-dependent:
\begin{equation}
    \mathcal{C}(t) = \{ x \in D \subset \mathbb{R}^n \mid h(x, p_i(t)) \geq 0\}.
\end{equation}

Adapting the standard CBF condition~\eqref{eq:define_cbf} to this time-varying setting yields:
\begin{equation}
\resizebox{\columnwidth}{!}{$
    \sup\limits_{u \in U}\left[L_f h_i(x, p_i(t)) + L_g h_i(x, p_i(t)) u + \frac{\partial h_i(x, p_i(t))}{\partial t} \right] \geq -\alpha(h_i(x, p_i(t))),
    \label{eq:tv_cbf}
$}
\end{equation}
where $\frac{\partial h_i(x, p_i(t))}{\partial t}$ 
accounts for the dynamics of each obstacle $p_i$, ensuring that the robot’s state remains within the safe set even as obstacles move.

For each obstacle $p_i$, let $d_{s}(p_i,q) $ denote the signed distance value from the differentiable representation~\eqref{eq:def_signed_distance} in the task space.
we construct the corresponding SDF-TVCBF constraint as $h^{s}_i(x,p_i(t))=d_s(p_i,q)$.
A notable benefit of our SDF-TVCBF is that they allow parallelized inference so that the distance computation scales efficiently even as the number of obstacles increases and is agnostic to the obstacle's geometries.

Assuming the robot's initial state satisfies $q(0) \in \mathcal{C}(0)$, we have $d_s(p_i(t), q(t)) > 0$ at $t=0$.
In static environments, ensuring that $\frac{\partial h(x)}{\partial x} \neq 0$ on the boundary of the safe set \(\mathcal{C}\) allows the robot to maintain a strictly safe distance from obstacles.
In dynamic environments, however, this condition becomes more stringent.
When $L_g h^{s}_i (x, t) = \frac{\partial d_s(p_i, q)}{\partial q} = 0$, it means that the robot is neither aware of the dynamics of the obstacle $p_i$ nor can react to it.
The gradient can become positive activating the TVCBF constraints again during a task as obstacles move, but the robot may be too close to the obstacle and lose the chance to avoid it in advance.
Numerous existing methods, including those using geometric primitives, polytopes, and learning-based models,
fall into this category~\cite{dai2023safe,koptev2022neural,koptev2024reactive,singletary2022safety}.
On the contrary, our second approach builds on top of the configuration distance and the gradient condition is guaranteed.
It is denoted as CDF-TVCBF and $h^{c}_i(x,t) = d_c(p_i,q)$.
Unlike CDF-TVCBF, which remains valid when the obstacle is within task space, SDF-TVCBF can retain its validity even when the obstacle is outside the task space, as long as the gradient exists, thereby maintaining the meaning of constraint \eqref{eq:tv_cbf}.

The first term in~\eqref{eq:tv_cbf}, $L_{f}h^{s}_i(x, p_i(t))$ remains zero.
The second term $L_g h_i(x,p_i(t)) = \frac{\partial h_i(x,p_i(t))}{\partial x} \equiv \frac{\partial d(p_i, q)}{\partial q}$
depends on the choice of the differentiable distance field.
It explains how each joint influences the distance-to-collision of the obstacle $p_i$.
When the distance field is $d_s(p_i,q)$, $\frac{\partial d_s(p_i,q)}{\partial q}$ measures the distance-to-collision in task space as shown in Fig.~\ref{fig:sdf_t_space}.
In contrast, $\frac{\partial d_c(p_i,q)}{\partial q}$ measures the distance-to-collision in configuration space.
Though the gradient$\frac{\partial d_s(p_i,q)}{\partial q}$ always exists, when it becomes zero, the SDF-TVCBF constraint is inactivated which means the manipulator is moving without considering the obstacle $p_i$.

For the third term $\frac{\partial h_i(x, p_i(t))}{\partial t} = \frac{\partial h_i(x, p_i(t))}{\partial p_i(t)} \frac{\partial p_i(t)}{\partial t} \equiv \frac{\partial d(p_i, q)}{\partial p_i} \frac{\partial p_i}{\partial t}$, it captures the dynamics of the obstacle $p_i$ in different spaces.
When the configuration distance is utilized,
$\frac{\partial h^c_i(x,p_i(t))}{\partial t} \equiv \frac{\partial d_c(p_i,q)}{\partial p_i} v_i$ measures the rate of change of the configuration-space distance to the obstacle $p_i$, influenced by the obstacle's velocity $v_i$, as shown in Fig.~\ref{fig:cdf_c_space}.
Therefore, we can design the CDF-TVCBF in a configuration space where the system geometry is a point.

\begin{myremark}
    The TVCBF constraint must not be relaxed, as it is essential for guaranteeing safety.
    However, it may not always be feasible, particularly if an obstacle approaches the base of the manipulator, which cannot be moved.
    Therefore, we assume that the moving obstacles do not approach the manipulator's base directly, and we consider that a theoretical solution exists, meaning the robot is agile enough, with sufficient physical limits, to avoid all obstacles.
\end{myremark}

The optimization problem~\eqref{eq:cbf_clf_qp_problem} optimizes only the control inputs and relaxation variables, allowing the direct integration of admissible control constraints into the QP formulation. For joint limits on the state variables \(q\), we define two functions to account for violations of physical limits:
\begin{equation}
    h_{\min}(q) = q - q_{\min}, \quad h_{\max}(q) = q_{\max} - q,
\end{equation}
which describe the boundary constraints as CBFs:
\begin{align}
    &L_f h_{\min}(x) + L_g h_{\min}(x) u \geq -\alpha_{\min}(h_{\min}(x)),\label{eq:physical_limits1}\\
    &L_f h_{\max}(x) + L_g h_{\max}(x) u \geq -\alpha_{\max}(h_{\max}(x)),
    \label{eq:physical_limits2}
\end{align}
where \(\alpha_{\max}\) and \(\alpha_{\min}\) are extended class \(\mathcal{K}\) functions.

The final optimization problem incorporating the CLF and CBF constraints, denoted as CDF-TVCBF-TVCLF, is given as 
\begin{align}
\label{eq:cdf-cbf_clf_qp_problem}
    u^{*}(x) = &\underset{u, \delta}{\operatorname{argmin}} 
    ~\frac{1}{2} u^{\top} R u + p \delta^2\\
    \text{s.t.}~ 
    &\eqref{eq:tv_clf}, \eqref{eq:tv_cbf}, \eqref{eq:physical_limits1},~\eqref{eq:physical_limits2} \nonumber\\
    & u \in U, \delta \in \mathbb{R}_{+}, \nonumber
\end{align}
where \( R \in \mathbb{R}^{n \times n} \) represents weights that prioritize control effort for each robotic joint, ensuring minimal effort. The parameter \( p \) controls the stabilization velocity and is particularly important in dynamic whole-body reaching tasks where exponential stabilization cannot be guaranteed, and the TVCLF is often non-monotone. 
Class functions \( \gamma \) and \( \alpha \) are chosen as simple linear scalar functions for computational efficiency. The TVCLF constraint guides the robot to generate dynamic whole-body reaching behaviors and is relaxed by \( \delta \) to handle transient deviations. The TVCBF constraint ensures that the robot avoids dynamic obstacles during operation.

\section{Numerical Simulations and Experiments}\label{sec:experiments}
\subsection{Implementation Details}
\begin{table}
    \caption{Setup of Simulation Parameters}
    \label{tab:simulation_params}
    \centering
    \begin{tabular}{l|l|l}
    \hline
    Notation & Meaning & Value     \\ \hline
    $\Delta t$ & Time step of simulation & $0.1 \, \si[per-mode=symbol]{\second}$ \\
    $l_1,l_2$ & The length of planar arm links & $2.0 \, \si[per-mode=symbol]{\metre}$\\
    $\epsilon_{\text{CBF}}$ & The safety margin for collision avoidance & $0.05 $\\
    $\epsilon_{\text{CLF}}$ & The tolerances for whole-body reaching & $0.02 \, \si[per-mode=symbol]{\radian}$\\
    $q_{\min}$ & Robot's minimum joint angles & $-\pi \, \si[per-mode=symbol]{\radian}$\\
    $q_{\max}$ & Robot's maximum joint angles & $\pi \, \si[per-mode=symbol]{\radian}$\\
    $\dot{q}_{\min}$ & Robot's   joint velocities & $-2.0 \, \si[per-mode=symbol]{\radian\per\second}$\\
    $\dot{q}_{\max}$ & Robot's maximum joint velocities & $2.0 \, \si[per-mode=symbol]{\radian\per\second}$\\
    $\gamma(\cdot)$ & The class functions for all TVCLFs & $1.0$\\
    $\alpha(\cdot)$ & The class functions for all TVCBFs & $1.0$ \\
    \hline     
    \end{tabular}%
\end{table}

In this section, we present the results of simulations and experiments on robotic systems to validate our approach.
To ensure safety, we incorporated a safety margin \( \epsilon_{\text{cbf}} \) in the TVCBF.
Similarly, a destination margin \( \epsilon_{\text{clf}} \) was included in the TVCLF to account for practical tolerances in whole-body reaching.
The simulation parameters for the robots and the optimization controller are summarized in Table~\ref{tab:simulation_params}.
The qpOASES solver~\cite{ferreau2014qpoases} was used to solve all QP optimization problems.
We conducted a series of simulations and experiments to evaluate our approach.
For additional results and details, please refer to the supplementary materials.

\subsection{2D Planar Arm Dynamic Collision avoidance}
\begin{figure*}[ht!]
    \centering
    \begin{subfigure}[t]{0.23\linewidth}
        \centering
        \includegraphics[width=\linewidth]{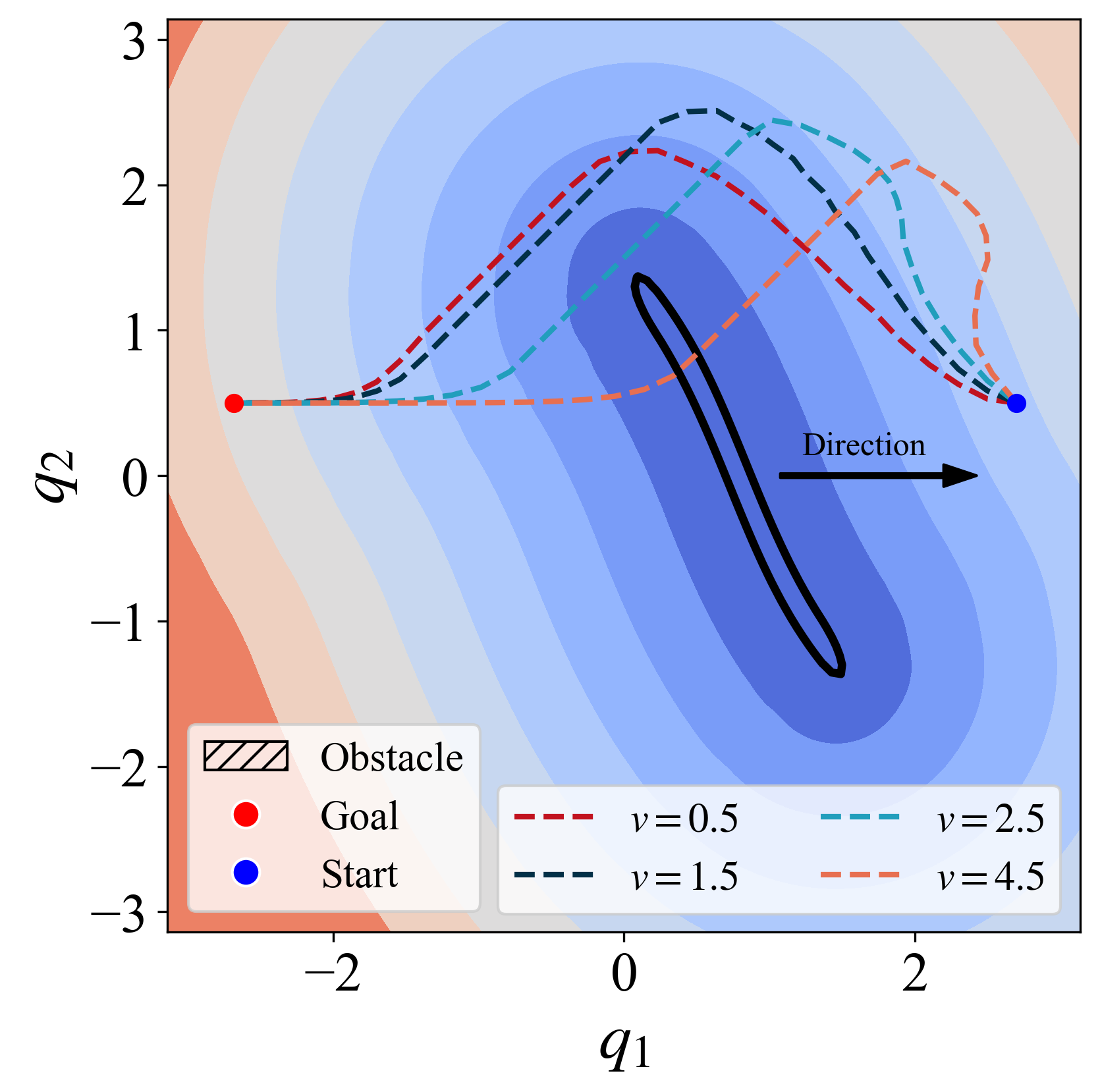}
        \caption{C-Space: CDF-TVCBF-QP}
        \label{fig:2d_MP_cdf_c_space}
    \end{subfigure}%
    \hspace{0.02\linewidth} 
    \begin{subfigure}[t]{0.23\linewidth}
        \centering
        \includegraphics[width=\linewidth]{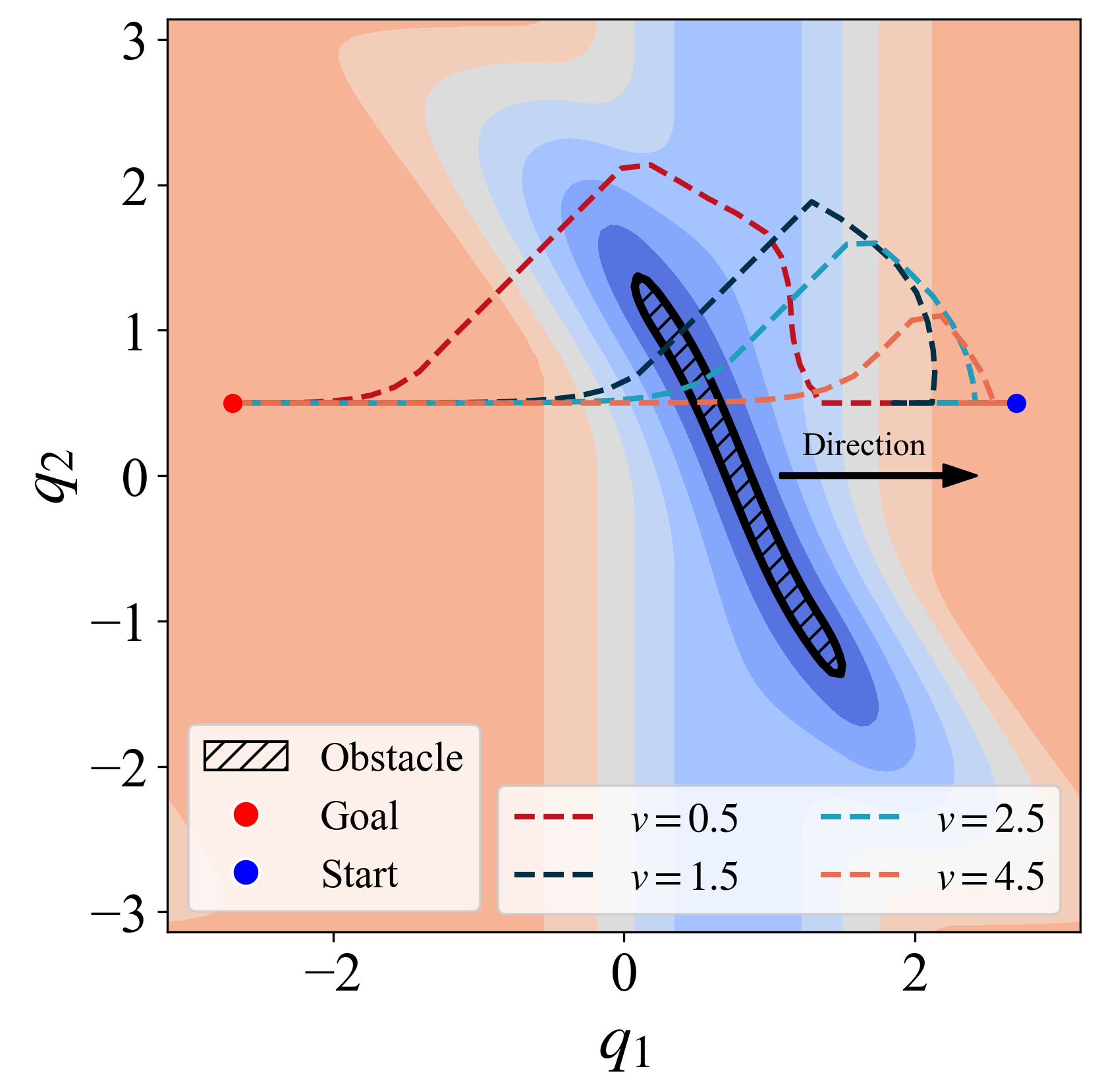}
        \caption{C-Space: SDF-TVCBF-QP}
        \label{fig:2d_MP_sdf_c_space}
    \end{subfigure}%
    \hspace{0.02\linewidth}
    \begin{subfigure}[t]{0.23\linewidth}
        \centering
        \includegraphics[width=\linewidth]{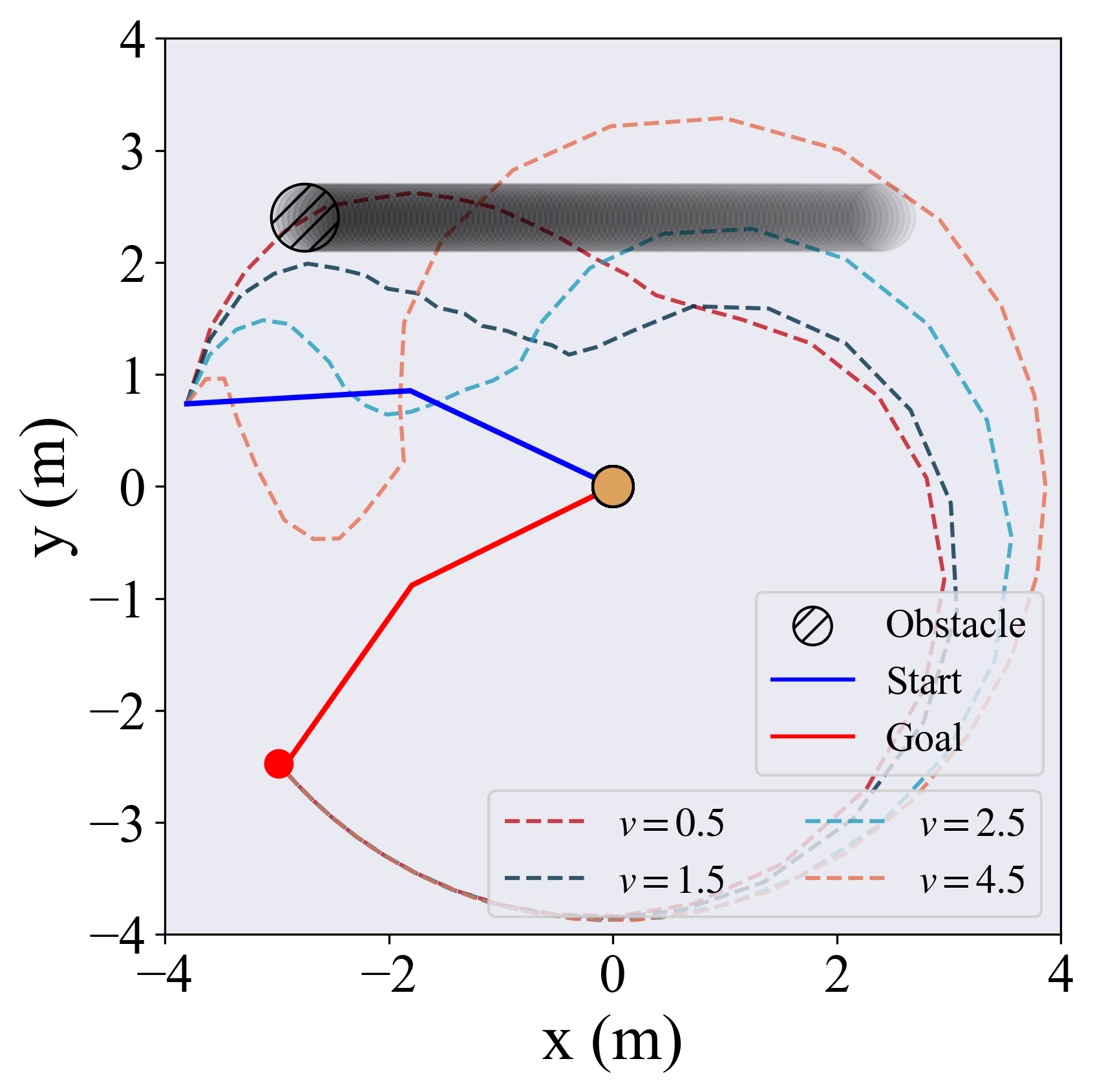}
        \caption{T-Space: CDF-TVCBF-QP}
        \label{fig:2d_MP_cdf_t_space}
    \end{subfigure}%
    \hspace{0.02\linewidth}
    \begin{subfigure}[t]{0.23\linewidth}
        \centering
        \includegraphics[width=\linewidth]{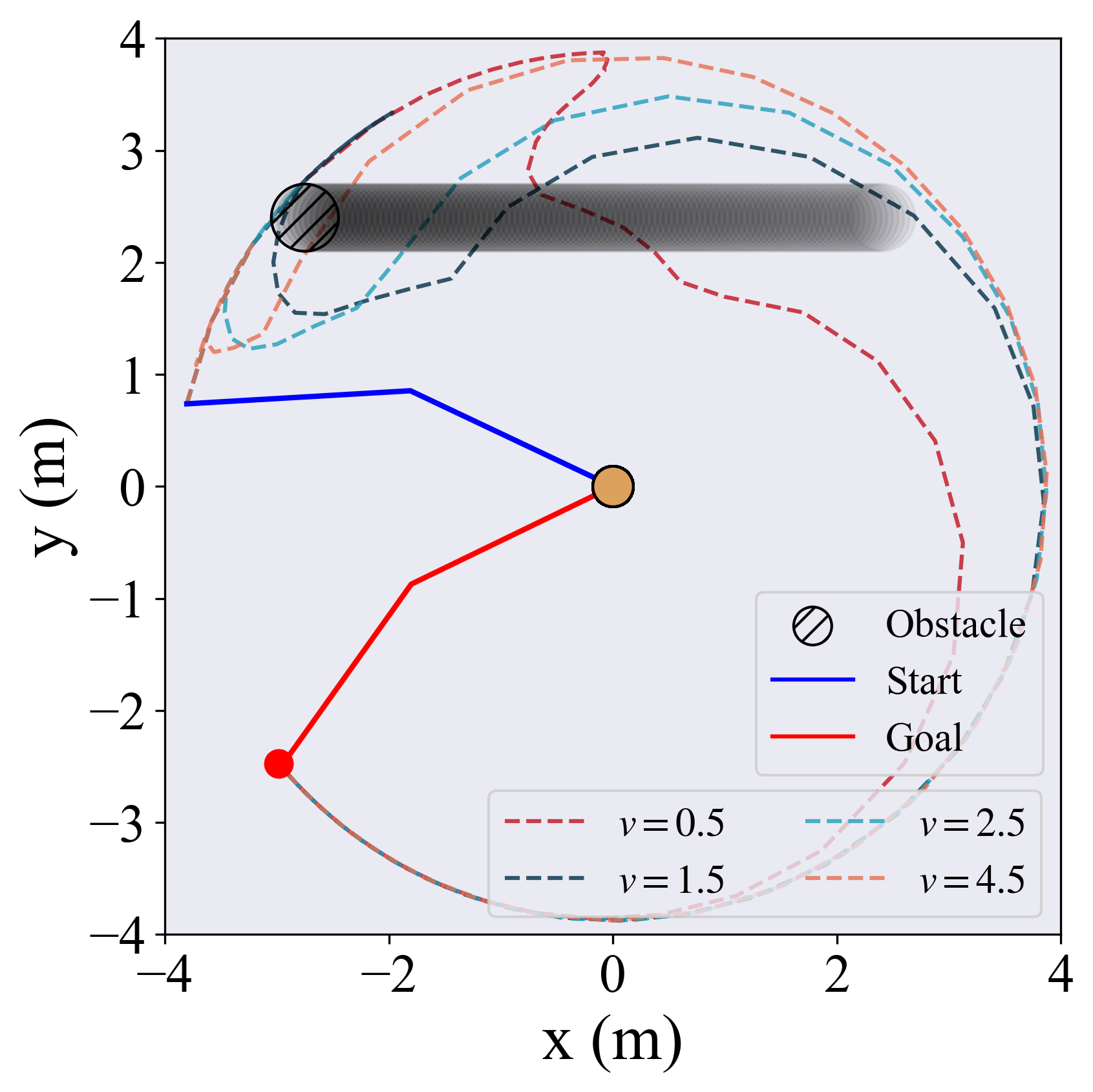}
        \caption{T-Space: SDF-TVCBF-QP}
        \label{fig:2d_MP_sdf_t_space}
    \end{subfigure}
    
    \caption{Illustration of the CDF/SDF-based TVCBF formulation.
    Arrows indicate the obstacle's motion direction.
    In the task space, the trajectories of the end-effector and the first joint are depicted in distinct colors.}
    \label{fig:MP_2d_demonstration}
\end{figure*}

The first scenario demonstrates the effectiveness of our SDF-TVCBF and CDF-TVCBF formulations that consider the velocity of obstacles for safety-critical control in dynamic environments.
The planar arm's initial state is $(q_1(0),q_2(0)) = (2.5, 0.5)$ and the goal state is $ q_{\text{goal}} = (-2.7, 0.5)$.
A moving circular obstacle with a radius of $0.3$ is introduced, initially positioned at $\left( 2.4, -2.4 \right)$ and moving at four different velocities $(0.5, 1.5, 2.5, 4.5) \, \si[per-mode=symbol]{\metre\per\second}$ to evaluate the effect of the obstacles dynamics.

As shown in Fig.~\ref{fig:2d_MP_cdf_c_space}, the CDF-TVCBF-QP controller effectively guides the robot to avoid obstacles under various velocity conditions.
At a velocity of $0.5 \, \si[per-mode=symbol]{\metre\per\second}$, the planar arm moves toward the target while avoiding the obstacle when necessary.
As the obstacle velocity increases to $1.5 \, \si[per-mode=symbol]{\metre\per\second}$ and $2.5  \, \si[per-mode=symbol]{\metre\per\second}$, the controller primarily adjust the joint angle $q_1$.
At the highest velocity of $4.5 \, \si[per-mode=symbol]{\metre\per\second}$, the controller ensures safety by moving the first link backward, enabling the robot to avoid the highly dynamic obstacle.
Additionally, the uniformly distributed level sets of the CDF in the configuration space confirm the existence of non-zero gradients, which is crucial for effective control. 
Fig.~\ref{fig:2d_MP_cdf_t_space} further illustrates safety-aware behaviors in the trajectories of both the end-effector and the first joint.

In contrast, the SDF-TVCBF constraint exhibits different safety-aware behaviors.
Unlike the CDF-TVCBF constraint, which proactively adjusts two joints to avoid the obstacle, the SDF-TVCBF keeps $q_2$ largely unchanged initially, relying solely on $q_1$ for obstacle avoidance.
This occurs because the gradient information $\frac{\partial h^{s}_i(x,t)}{\partial q}$ does not effectively guide the motion of $q_2$.
At the initial stage, the closest point to the obstacle lies on the first link, resulting in a zero gradient with respect to $q_2$
As shown in Fig.~\ref{fig:2d_MP_sdf_c_space}, the level sets of the SDF reveal that gradients either become zero or align parallel to $q_1$.
In more general scenarios, when the closest point shifts to the manipulator's base, the gradient becomes zero and the manipulator is not able to react effectively to the moving obstacle.
From path length of end-effector trajectories shown in ~Fig.~\ref{fig:2d_MP_cdf_t_space} and~Fig.~\ref{fig:2d_MP_sdf_t_space}, we can also see that complete gradient information would help converge faster.

\subsection{Planar Arm Dynamic and collision-free whole-body reaching}
\begin{figure*}[ht!]
    \centering
    \begin{subfigure}[t]{0.23\linewidth}
        \centering
        \includegraphics[width=\linewidth]{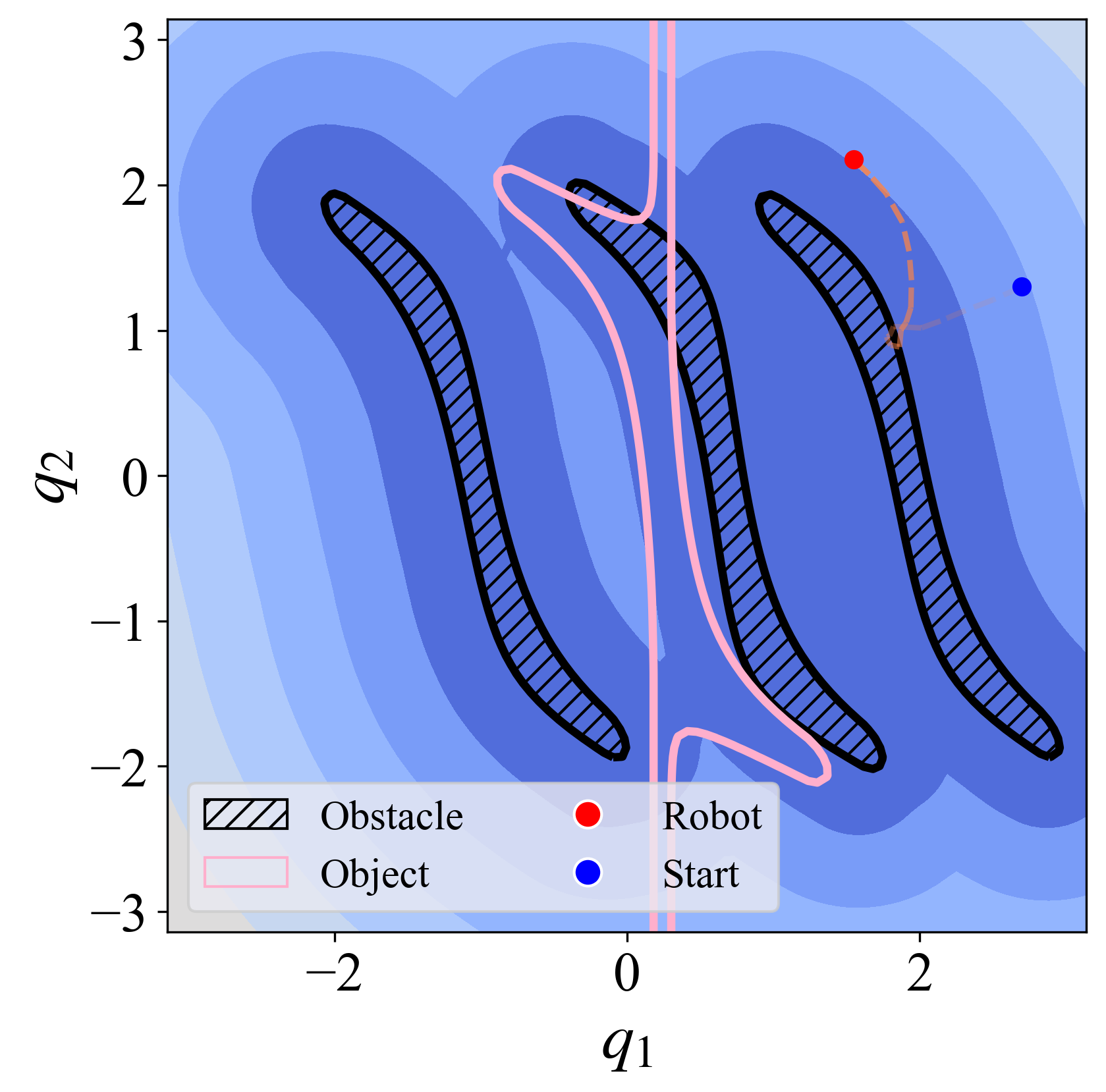}
        \caption{C-Space}
        \label{fig:2d_whole_body_reaching_c_space}
    \end{subfigure}%
    \hspace{0.02\linewidth} 
    \begin{subfigure}[t]{0.23\linewidth}
        \centering
        \includegraphics[width=\linewidth]{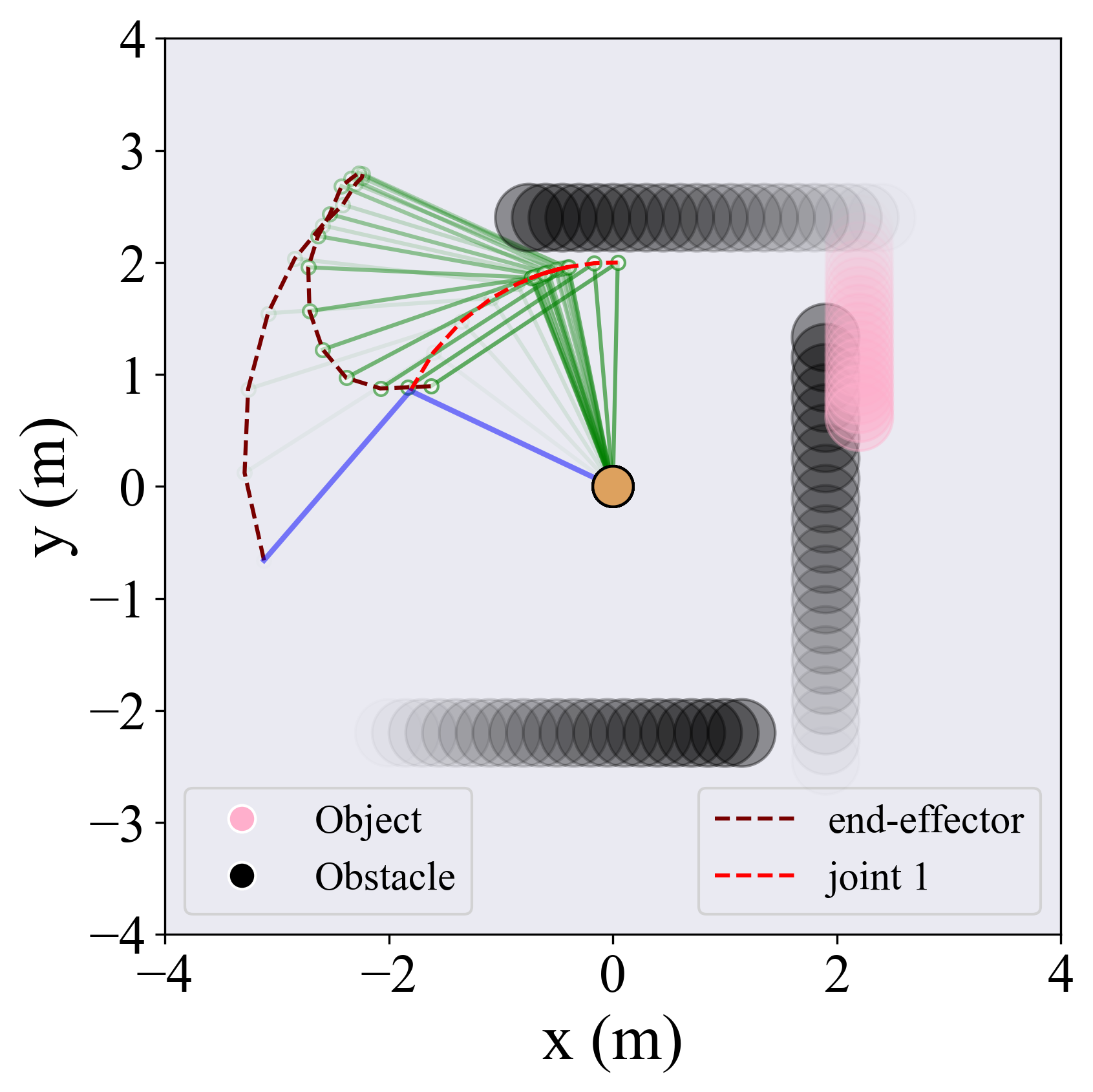}
        \caption{T-Space}
        \label{fig:2d_whole_body_reaching_t_space}
    \end{subfigure}%
    \hspace{0.02\linewidth}
    \begin{subfigure}[t]{0.23\linewidth}
        \centering
        \includegraphics[width=\linewidth]{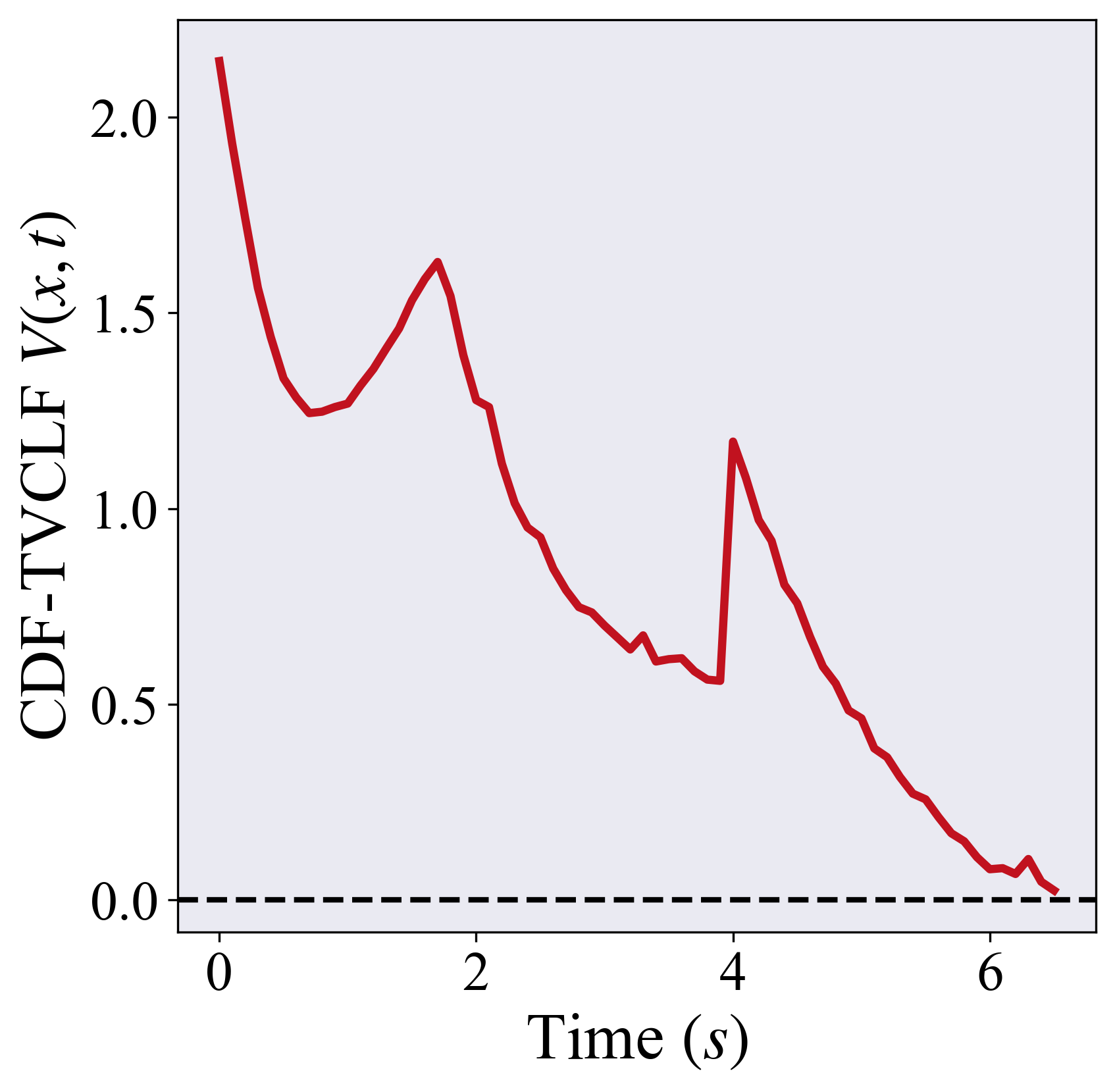}
        \caption{TVCLF values}
        \label{fig:2d_whole_body_reaching_clf}
    \end{subfigure}%
    \hspace{0.02\linewidth}
    \begin{subfigure}[t]{0.23\linewidth}
        \centering
        \includegraphics[width=\linewidth]{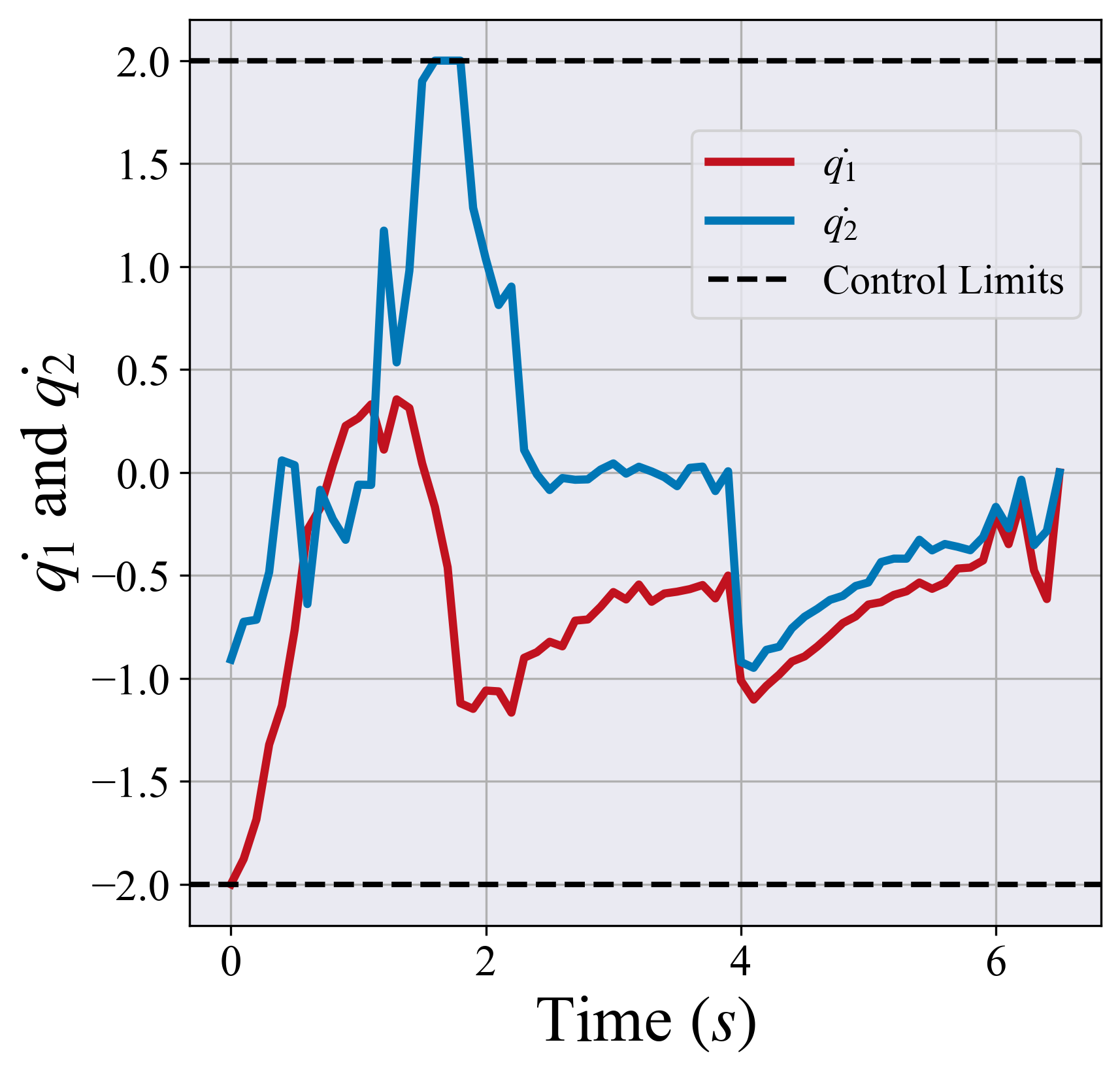}
        \caption{Control variables}
        \label{fig:2d_whole_body_reaching_ut}
    \end{subfigure}
    \caption{The illustration of collision-free and whole-body reaching in dynamic environments.
    The first two figures are a snapshot at $t=2.2 \, \si[per-mode=symbol]{\second}$.
    The values of TVCLF and control variables are full-time horizons to show the quantitative details.}
    \label{fig:2d_whole_body_reaching}
\end{figure*}
This section demonstrates the efficacy of the CDF-TVCBF-TVCLF-QP formulation in a dynamic reaching task.
While our method is agnostic to object geometry, a circular object with a radius of $0.3 \, \si[per-mode=symbol]{\metre}$ is used for simplicity.
The target is initially placed at $p_g = (2.2, 2.3)$ and moves downward with a velocity $v_g = (0.0, -0.8) \, \si[per-mode=symbol]{\metre\per\second}$.
Three obstacles, positioned at $p_1 = (1.9, -2.45)$, $p_2 = (2.4, 2.4)$, move with velocities $v_1 =(0.0, 1.8) \, \si[per-mode=symbol]{\metre\per\second}$ (upward), $v_2= (-1.5, 0.0) \, \si[per-mode=symbol]{\metre\per\second}$ (leftward), and $v_3=(1.5, 0.0) \, \si[per-mode=symbol]{\metre\per\second}$ (rightward).
The results are illustrated in Fig.~\ref{fig:2d_whole_body_reaching}. The planar arm progresses through four distinct stages to reach the target object:

$\textit{Stage 1: Initial Approach}$ ($t= 0.0 \, \si[per-mode=symbol]{\second}$ to $t = 0.8 \, \si[per-mode=symbol]{\second}$).
During this stage, the robot moves toward the target object while avoiding the slow-moving obstacle.
Joint velocities are initially negative, allowing the robot to approach the target while avoiding collisions. 
As shown in Fig.~\ref{fig:2d_whole_body_reaching_ut}, joint velocities $\dot{q}_1$ and $\dot{q}_2$ begin to increase toward positive values, indicating that the robot is preemptively planning to avoid future obstacles.
The TVCLF decreases, signifying progress toward the target.

$\textit{Stage 2: Safety Prioritization}$ ($t= 0.8 \, \si[per-mode=symbol]{\second}$ to $t = 1.7 \, \si[per-mode=symbol]{\second}$).
In this stage, the controller prioritizes safety over whole-body reaching, leading to an increase in the TVCLF as the robot temporarily moves away from the target.
Joint velocity $\dot{q}_1$ becomes positive, causing the arm to adjust its configuration to avoid collisions, as shown in Fig.~\ref{fig:2d_whole_body_reaching_t_space}.
Meanwhile, the target object continues moving downward, increasing the TVCLF.

$\textit{Stage 3: Gradual Reaching}$ ($t= 1.7 \, \si[per-mode=symbol]{\second}$ to $t = 3.9 \, \si[per-mode=symbol]{\second}$).
At this stage, the second obstacle overlaps with the target object, creating configurations that allow reaching but are not collision-free. 
The robot aims to reach the zero level set of the target object, as shown in Fig.\ref{fig:2d_whole_body_reaching_c_space}.
Since the first link is closer to the target, the controller attempts to reach the object using this link.
As demonstrated in Fig.\ref{fig:2d_whole_body_reaching_clf}, the TVCLF decreases, indicating progress toward the target.
Joint velocity $\dot{q}_1$ becomes negative, rotating the link clockwise to approach the target.
However, the setup intentionally places the target object further from the robot, limiting the available time window for the first link to complete the reach. By $t=3.9 \, \si[per-mode=symbol]{\second}$, this time window closes, making it impossible for the first link to reach the target.
This is reflected in a brief increase in the TVCLF (the cusp), indicating the need for a strategy adjustment.

$\textit{Stage 4: Switching Strategy}$ ($t= 3.9 \, \si[per-mode=symbol]{\second}$ to $t = 6.7 \, \si[per-mode=symbol]{\second}$).
After the first link's time window closes at $t= 3.9 \, \si[per-mode=symbol]{\second}$, the CDF-TVCBF-TVCLF-QP controller switches its strategy to use the second link to reach the target.
At this point, the gradient of the TVCLF with respect to $q_2$ becomes active, enabling the robot to adjust its configuration effectively.
Joint velocity $q_2$ becomes negative, gradually moving the second link closer to the target object.
The TVCLF decreases steadily between $t=3.9 \, \si[per-mode=symbol]{\second}$ and $t=6.7 \, \si[per-mode=symbol]{\second}$.
As the robot approaches the target, joint velocities slow, and the planar arm successfully reaches the target object with the second link.

We observe that relaxing the TVCLF constraint is essential for two reasons. First, the controller must balance safety and stabilization under dynamic conditions. Second, as the target object moves, the TVCLF may increase, requiring the robot to adaptively decide which body part to use for reaching, ensuring effective and collision-free motion generation.

\subsection{Performance Evaluation}
\begin{table*}[]
\caption{Performance benchmark: motion planning in highly dynamic environments.}
\label{tab:benchmark}
\resizebox{\textwidth}{!}{
\begin{tabular}{l|ccccccccc|ccccccccc|ccc}
\hline
\multicolumn{1}{c|}{\multirow{3}{*}{Methods}} & \multicolumn{9}{c|}{Time to Reach (s)}                                                                   & \multicolumn{9}{c|}{Path Length (m)}                                                                         & \multicolumn{3}{c}{Success Rate} \\ \cline{2-22} 
\multicolumn{1}{c|}{}                         & \multicolumn{3}{c|}{S1}               & \multicolumn{3}{c|}{S2}                & \multicolumn{3}{c|}{S3} & \multicolumn{3}{c|}{S1}                  & \multicolumn{3}{c|}{S2}                 & \multicolumn{3}{c|}{S3} & S1        & S2        & S3       \\ \cline{2-22} 
\multicolumn{1}{c|}{}                         & min & max & \multicolumn{1}{c|}{avg}  & min & max  & \multicolumn{1}{c|}{avg}  & min   & max    & avg    & min  & max   & \multicolumn{1}{c|}{avg}  & min  & max  & \multicolumn{1}{c|}{avg}  & min    & max    & avg   & -         & -         & -        \\ \hline
CDF-TVCBF-QP (ours)                           & 2.9 & 3.8 & \multicolumn{1}{c|}{3.2}  & 3.2 & 4.5  & \multicolumn{1}{c|}{3.8}  & 3.0   & 6.4    & 4.4    & 6.12 & 7.50  & \multicolumn{1}{c|}{6.72} & 7.48 & 8.75 & \multicolumn{1}{c|}{7.47} & 6.46   & 10.43  & 8.25  & 1.00      & 1.00      & 1.00     \\ \cline{1-1}
SDF-TVCBF-QP (ours)                           & 3.5 & 4.8 & \multicolumn{1}{c|}{3.9}  & 3.4 & 4.5  & \multicolumn{1}{c|}{3.9}  & 3.1   & 6.4    & 4.8    & 6.22 & 7.71  & \multicolumn{1}{c|}{6.64} & 6.30 & 8.03 & \multicolumn{1}{c|}{7.06} & 6.38   & 12.37  & 8.52  & 1.00      & 1.00      & 1.00     \\ \cline{1-1}
SDF-CBF-QP (\cite{singletary2022safety})                             & 2.9 & 2.9 & \multicolumn{1}{c|}{19.8} & -   & -    & \multicolumn{1}{c|}{-}    & -     & -      & -      & 6.13 & 6.13  & \multicolumn{1}{c|}{6.13} & -    & -    & \multicolumn{1}{c|}{-}    & -      & -      & -     & 0.01      & 0.00      & 0.00     \\ \cline{1-1}
SDF-SH-QP (\cite{li2024representing}, \cite{koptev2022neural})                              & 3.1 & 4.9 & \multicolumn{1}{c|}{8.7}  & 3.2 & 4.4  & \multicolumn{1}{c|}{13.2}  & 3.0   & 6.3    & 9.5    & 6.73 & 9.52  & \multicolumn{1}{c|}{7.58} & 6.66 & 8.69 & \multicolumn{1}{c|}{7.41} & 6.63   & 10.43  & 8.85  & 0.69      & 0.41      & 0.68     \\ \cline{1-1}
CDF-SH-QP (\cite{li2024configuration})                              & 3.1 & 4.7 & \multicolumn{1}{c|}{6.9}  & 3.1 & 4.4  & \multicolumn{1}{c|}{12.6} & 3.0   & 6.3    & 7.2    & 6.54 & 10.04 & \multicolumn{1}{c|}{7.48} & 6.82 & 9.63 & \multicolumn{1}{c|}{7.62} & 6.68   & 13.62  & 9.60  & 0.79      & 0.45      & 0.82     \\ \hline
\multicolumn{1}{c|}{7D}                       & \multicolumn{3}{c|}{C1}               & \multicolumn{3}{c|}{C2}                & \multicolumn{3}{c|}{C3} & \multicolumn{3}{c|}{C1}                  & \multicolumn{3}{c|}{C2}                 & \multicolumn{3}{c|}{C3} & C1        & C2        & C3       \\ \hline
CDF-TVCBF-QP (ours)                           & 2.8 & 4.0 & \multicolumn{1}{c|}{3.1}  & 3.5 & 15.8 & \multicolumn{1}{c|}{12.3} & 3.3   & 10.3   & 7.1    & 2.64 & 4.09  & \multicolumn{1}{c|}{3.24} & 2.06 & 7.19 & \multicolumn{1}{c|}{4.43} & 3.67   & 6.84   & 4.83  & 1.00      & 0.66      & 0.92     \\
SDF-TVCBF-QP (ours)                           & 2.8 & 4.3 & \multicolumn{1}{c|}{3.3}  & 3.4 & 10.5 & \multicolumn{1}{c|}{14.0} & 3.7   & 13.0   & 9.6    & 2.78 & 4.35  & \multicolumn{1}{c|}{3.51} & 2.29 & 4.80 & \multicolumn{1}{c|}{2.77} & 4.21   & 10.04  & 5.09  & 1.00      & 0.42      & 0.81     \\
SDF-CBF-QP (\cite{singletary2022safety})                             & 2.8 & 3.1 & \multicolumn{1}{c|}{9.3}  & -   & -    & \multicolumn{1}{c|}{-}    & 3.0   & 4.2    & 13.9   & 2.38 & 2.16  & \multicolumn{1}{c|}{2.63} & -    & -    & \multicolumn{1}{c|}{-}    & 3.53   & 4.03   & 3.73  & 0.62      & 0.00      & 0.37     \\
SDF-SH-QP (\cite{li2024representing}, \cite{koptev2022neural})                              & 2.5 & 3.4 & \multicolumn{1}{c|}{9.7}  & -   & -    & \multicolumn{1}{c|}{-}    & 2.6   & 6.7    & 16.2   & 2.09 & 3.97  & \multicolumn{1}{c|}{2.78} & -    & -    & \multicolumn{1}{c|}{-}    & 3.75   & 6.44   & 4.62  & 0.63      & 0.00      & 0.24     \\
CDF-SH-QP (\cite{li2024configuration})                              & 2.5 & 2.9 & \multicolumn{1}{c|}{9.3}  & -   & -    & \multicolumn{1}{c|}{-}    & 2.5   & 2.8    & 15.8   & 2.07 & 3.49  & \multicolumn{1}{c|}{2.68} & -    & -    & \multicolumn{1}{c|}{-}    & 3.07   & 3.75   & 3.23  & 0.59      & 0.00      & 0.24     \\ \hline
\end{tabular}
}
\end{table*}

\begin{figure}[t!]
    \centering
    \begin{subfigure}[t]{0.48\linewidth}
        \centering
        \includegraphics[width=\linewidth]{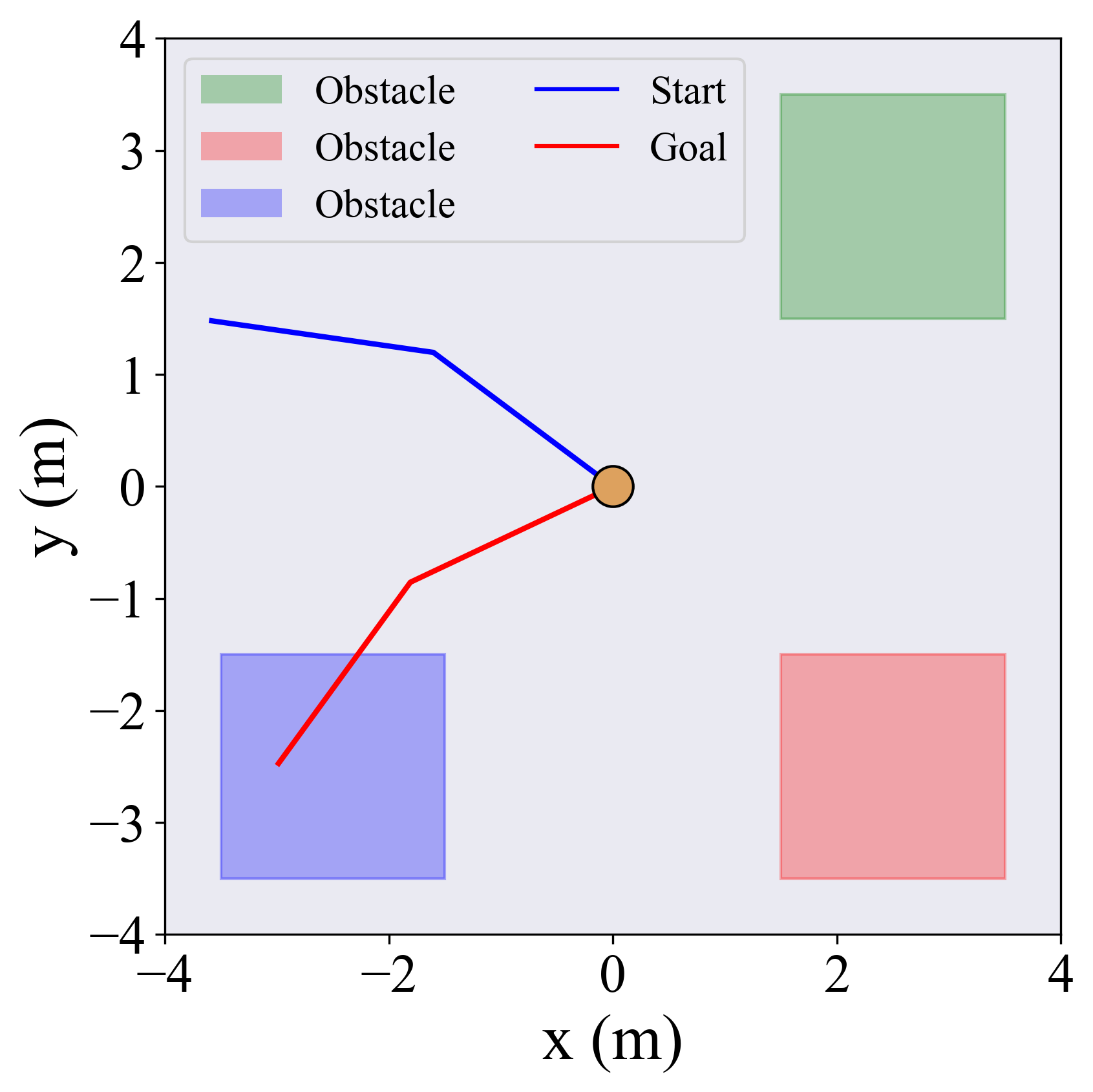}
        \caption{2D benchmark setup}
        \label{fig:2d_bench}
    \end{subfigure}%
    \hspace{0.02\linewidth} 
    \begin{subfigure}[t]{0.48\linewidth}
        \centering
        \includegraphics[width=\linewidth]{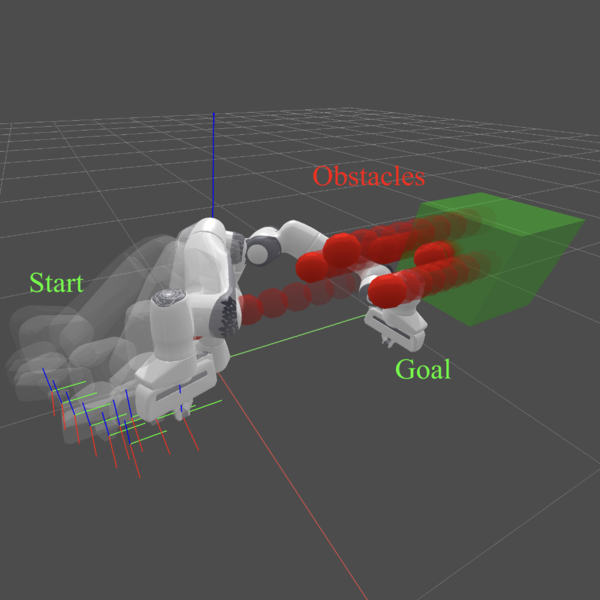}
        \caption{7D benchmark setup}
        \label{fig:7d_bench}
    \end{subfigure}
    \caption{The 2D and 7D setup for benchmarking.
    Dynamic obstacles are randomly generated from the 2D and 7D box-shape region.}
    \label{fig:bench_setup}
\end{figure}

\begin{figure*}[ht!]
    \centering
    \begin{subfigure}[t]{0.19\linewidth}
        \centering
        \includegraphics[width=\linewidth]{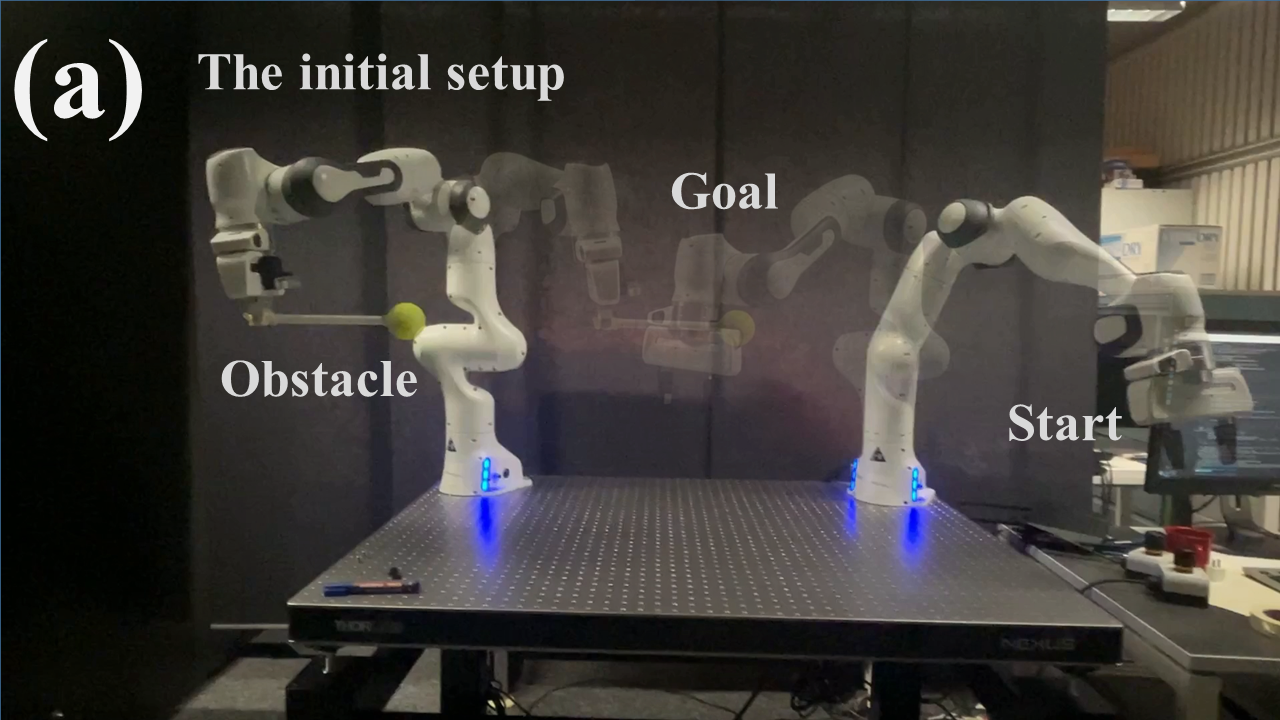}
    \end{subfigure}%
    \hspace{0.005\linewidth}
    \begin{subfigure}[t]{0.19\linewidth}
        \centering
        \includegraphics[width=\linewidth]{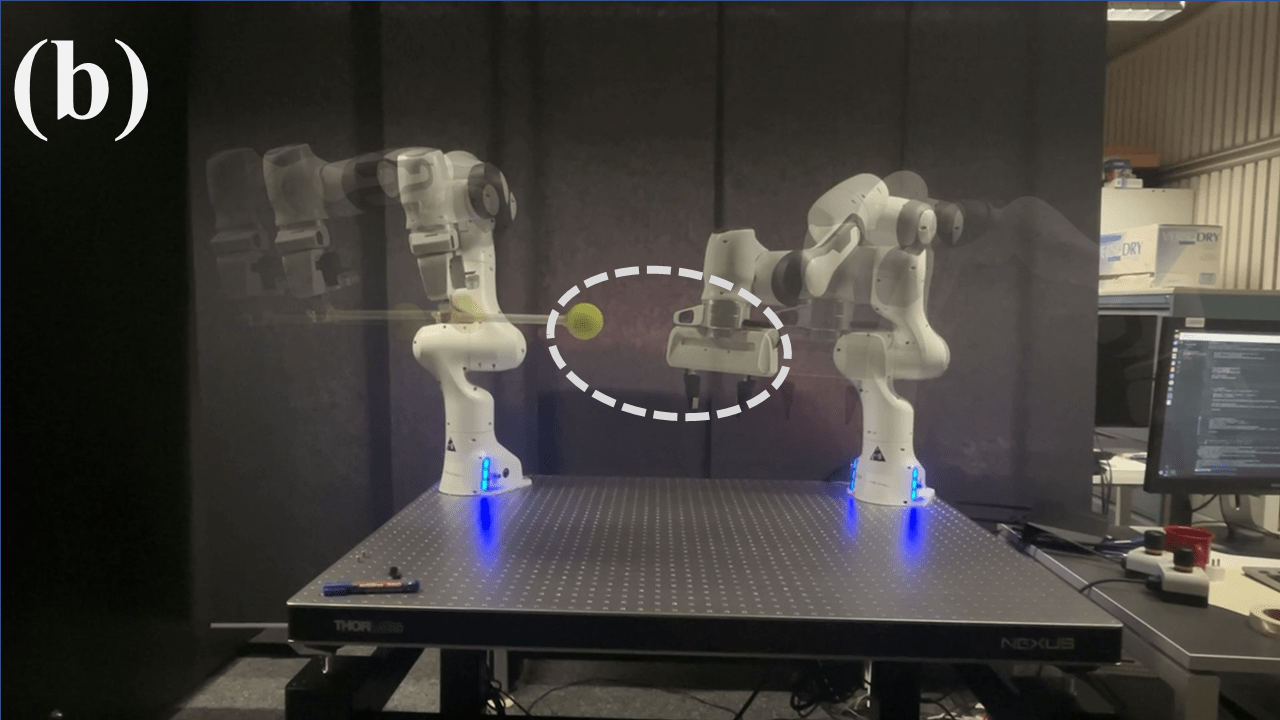}
    \end{subfigure}%
    \hspace{0.005\linewidth}
    \begin{subfigure}[t]{0.19\linewidth}
        \centering
        \includegraphics[width=\linewidth]{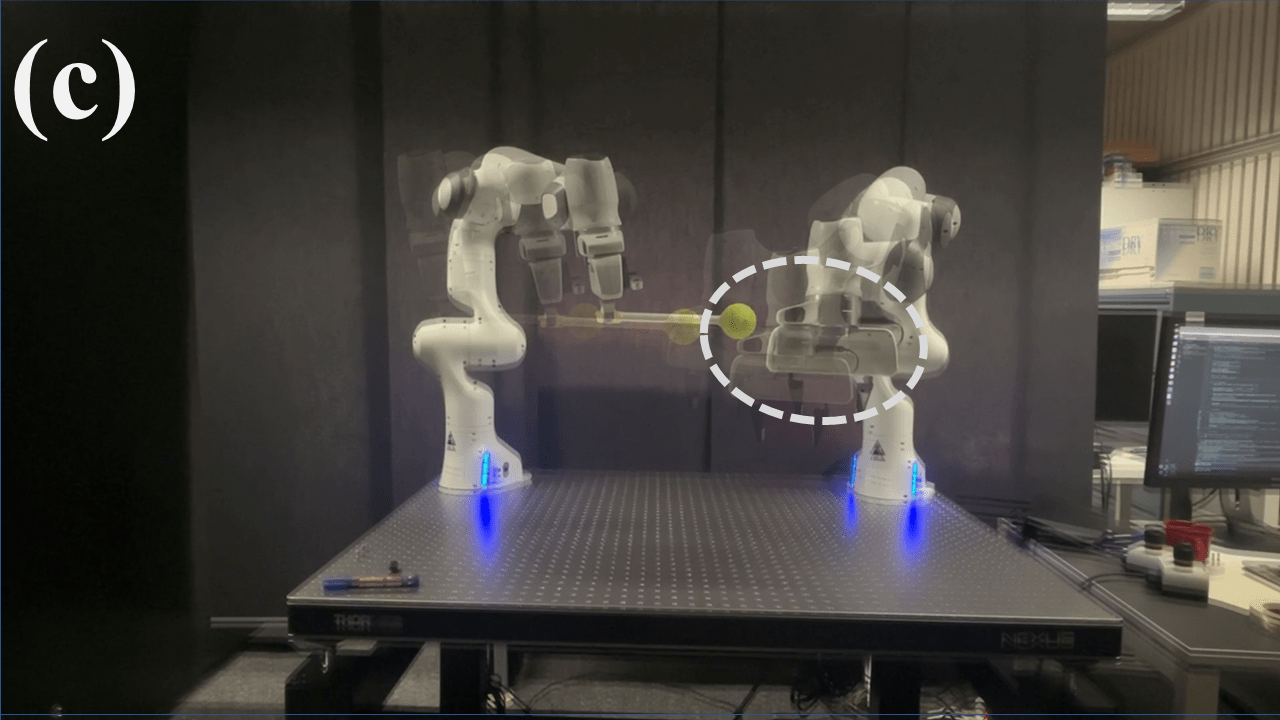}
    \end{subfigure}%
    \hspace{0.005\linewidth}
    \begin{subfigure}[t]{0.19\linewidth}
        \centering
        \includegraphics[width=\linewidth]{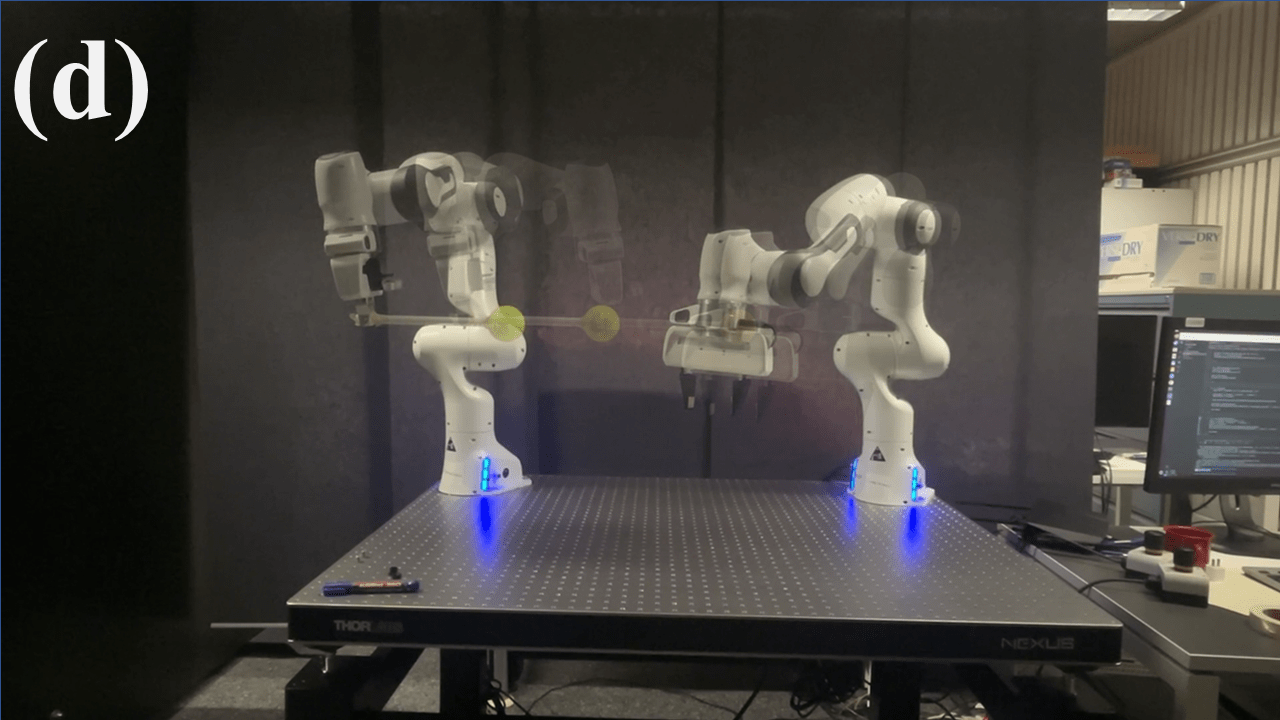}
    \end{subfigure}%
    \hspace{0.005\linewidth}
    \begin{subfigure}[t]{0.19\linewidth}
        \centering
        \includegraphics[width=\linewidth]{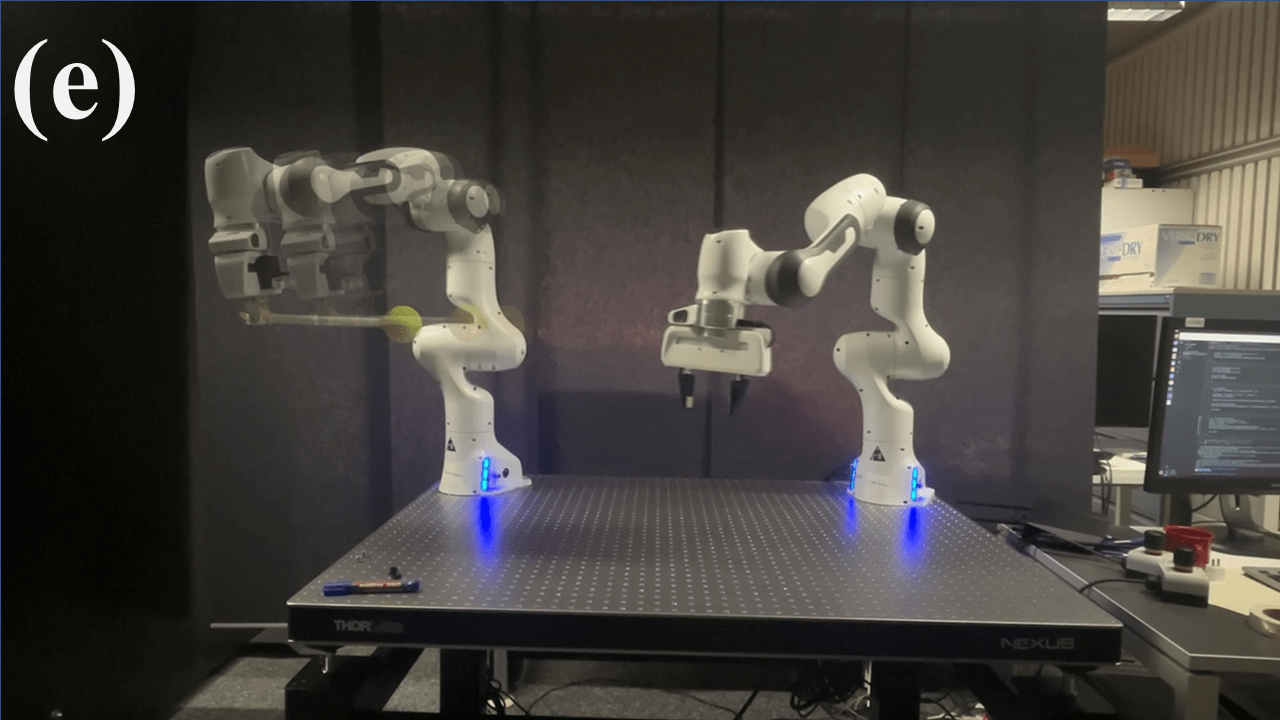}
    \end{subfigure}
    
    \vspace{0.01\linewidth}
    \begin{subfigure}[t]{0.19\linewidth}
        \centering
        \includegraphics[width=\linewidth]{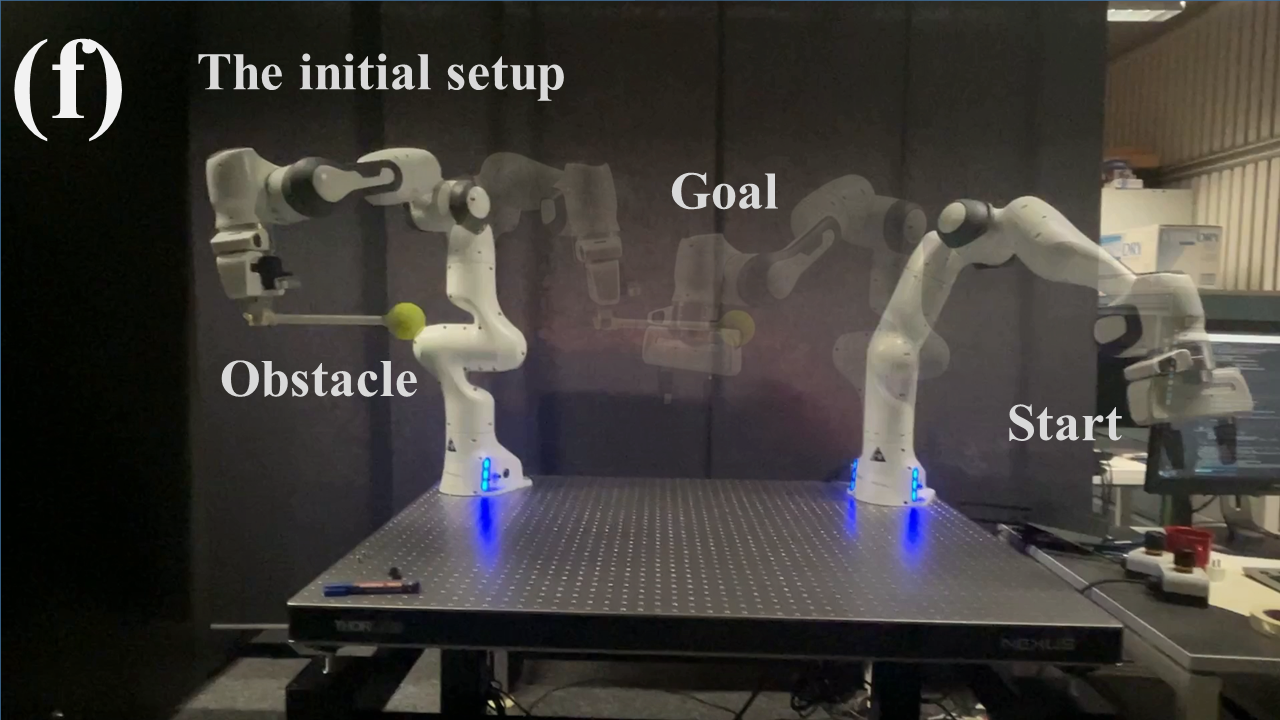}
    \end{subfigure}%
    \hspace{0.005\linewidth}
    \begin{subfigure}[t]{0.19\linewidth}
        \centering
        \includegraphics[width=\linewidth]{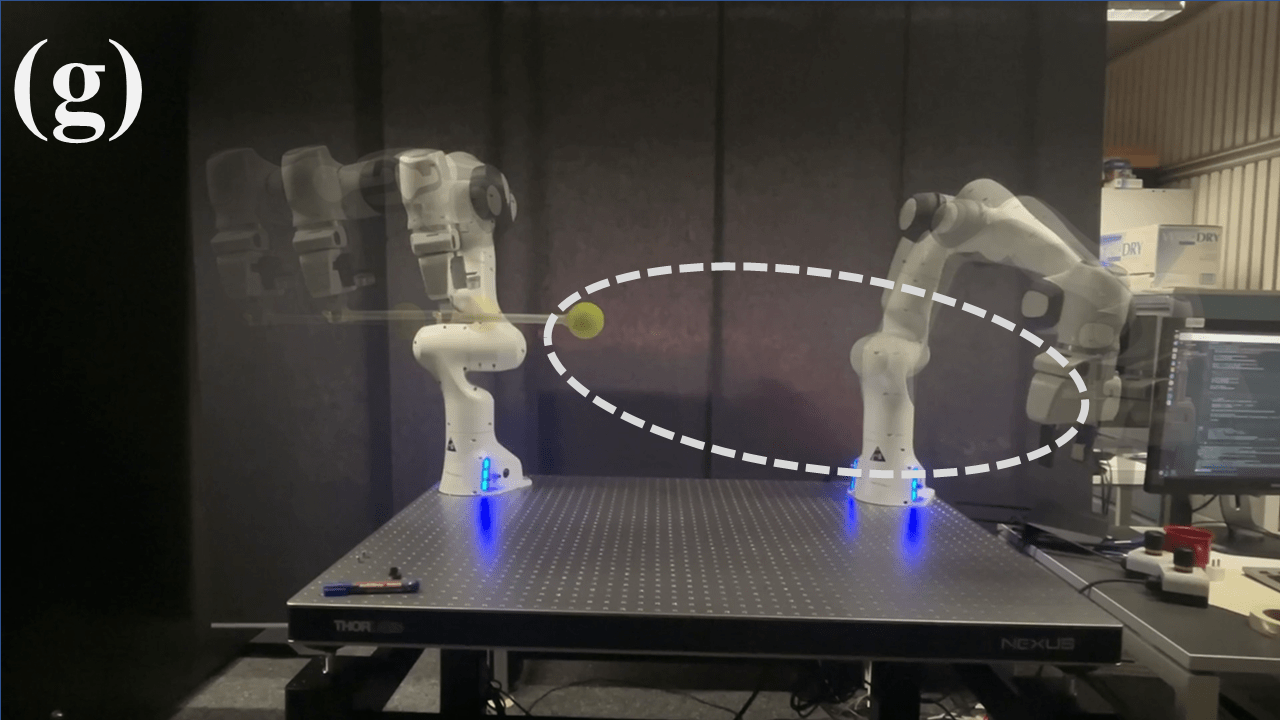}
    \end{subfigure}%
    \hspace{0.005\linewidth}
    \begin{subfigure}[t]{0.19\linewidth}
        \centering
        \includegraphics[width=\linewidth]{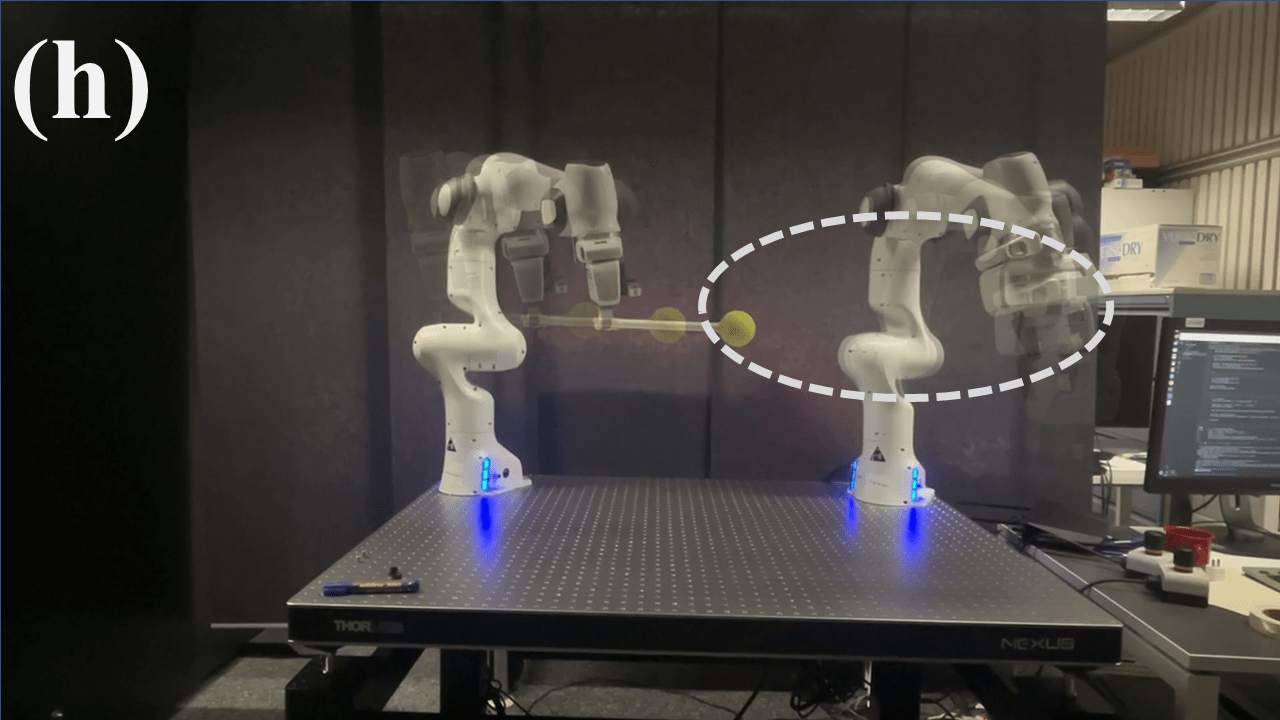}
    \end{subfigure}%
    \hspace{0.005\linewidth}
    \begin{subfigure}[t]{0.19\linewidth}
        \centering
        \includegraphics[width=\linewidth]{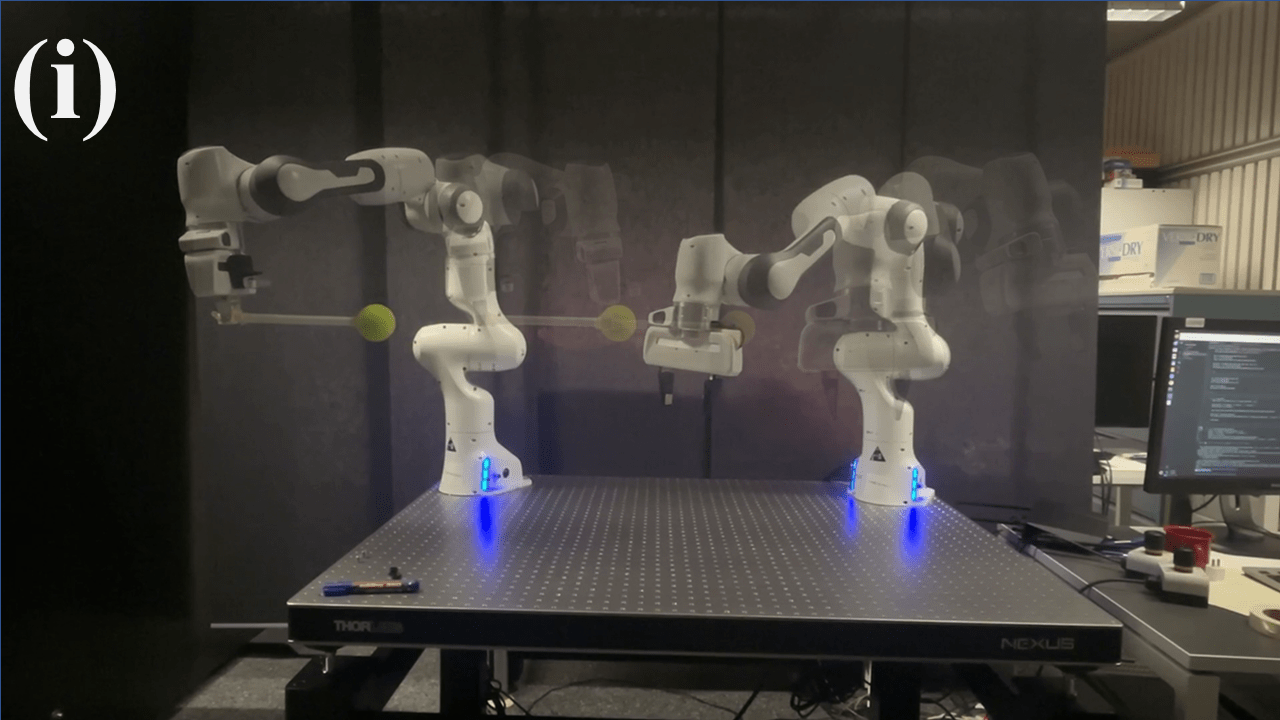}
    \end{subfigure}%
    \hspace{0.005\linewidth}
    \begin{subfigure}[t]{0.19\linewidth}
        \centering
        \includegraphics[width=\linewidth]{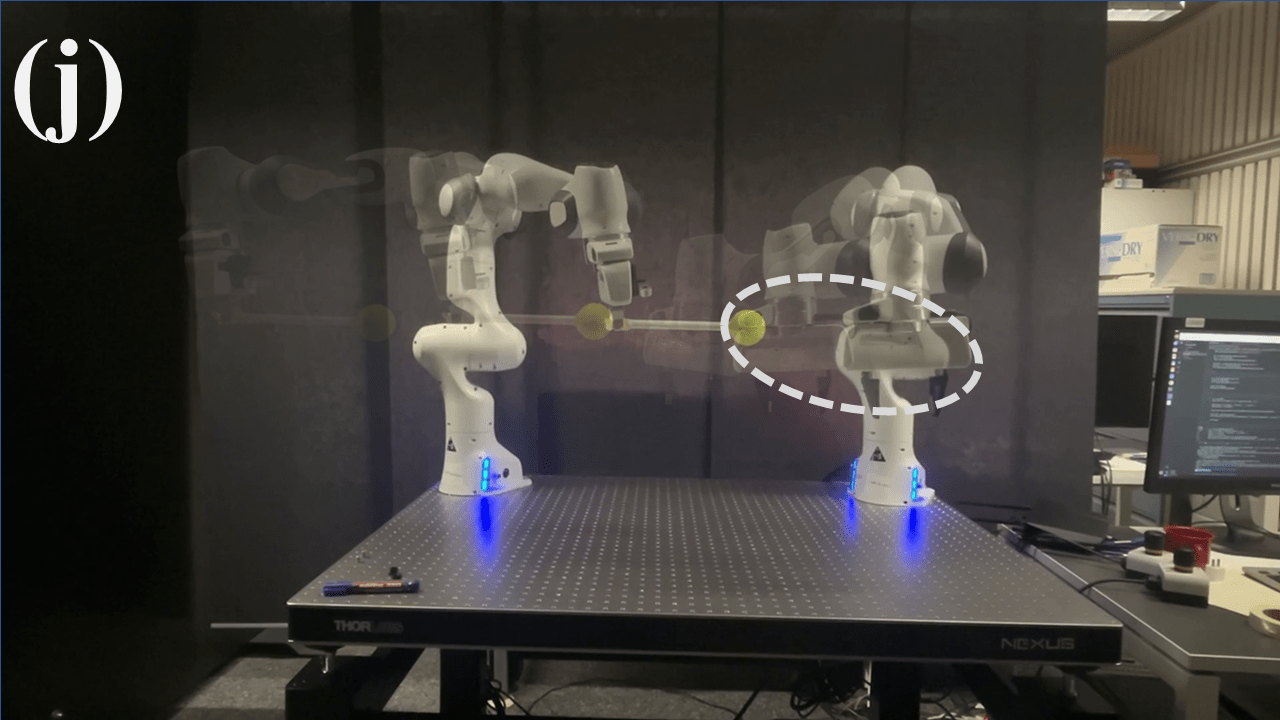}
        \label{fig:exp_j}
    \end{subfigure}%
    \caption{CDF-TVCBF-QP: snapshots of the collision-avoidance task with different velocities of obstacles.
    (a)~The obstacle moves slowly at $0.05~\si[per-mode=symbol]{\metre\per\second}$. 
    (b)~The robot reaches the goal early, as it is aware of the obstacle's dynamics.
    (c)~Due to the slow movement of the obstacle, the robot arm stays nearby.
    (d)-(e)~The robot returns to the goal and maintains its position as the obstacle moves far away.
    (f)~The obstacle moves slowly at $0.15~\si[per-mode=symbol]{\metre\per\second}$. 
    (g)-(h)~The robot behaves in a safety-aware manner and stays farther away due to the high velocity of the obstacle.
    (i)-(j)~The robot arm reaches the goal while prioritizing safety.}
    \label{fig:robot_experiments}
\end{figure*}

To quantitatively evaluate our method, we compare its performance with a set of baselines.
The first two baselines construct safety constraints using separation hyperplanes (SH) based on gradient information, denoted as SDF-SH-QP~\cite{koptev2022neural}
and CDF-SH-QP~\cite{li2024configuration}.
The third baseline is denoted as SDF-CBF-QP proposed in~\cite{singletary2022safety}.
This method formulates CBFs using signed distances computed by GJK and EPA algorithms.
Since these baselines do not include CLF constraints, we also test simplified versions of our approach, denoted as CDF-TVCBF-QP and SDF-TVCBF-QP, with the CLF constraint removed.
We use the following metrics to assess controller performance:
\begin{enumerate}
    \item \textbf{Success Rate}: The percentage of trials where the controller successfully reaches the goal while avoiding obstacles.
    \item \textbf{Time to Reach}: The time taken to reach the goal in seconds.
    If the controller fails, the maximum allowable time $20 \, \si[per-mode=symbol]{\second}$ is recorded.
    \item \textbf{Path Length}: The $\ell^2$ norm of the whole-body trajectory length in meters.
\end{enumerate}

We conduct experiments across three 2D scenarios (S1, S2, and S3) and three 7D scenarios (C1, C2, and C3), each comprising 100 randomly generated setups.
These experiments evaluate the controller performance in dynamic environments using metrics that capture both effectiveness (success rate) and efficiency (time-to-reach and path length).
In scenario S1, a single moving obstacle is randomly generated within the region $x\in \left[1.5, 3.5 \right] \m$ and $y\in \left[1.5, 3.5 \right] \m$.
The initial and goal states of the planar arm are $q(0) = (2.5, 0.5) \rad$ and $q_g = (-2.7, 0.5) \rad$.
The obstacle velocity is randomly selected within $(-4, -2) \mps $ along the x-axis.
In scenario S2, a second obstacle is added, randomly generated within $x \in \left [ 1.5, 3.0\right ] \m$ and $y \in \left[-3.0, -1.5 \right] \m$.
The velocity range of the obstacle is increased, $v_1 \in \left[3.0, 5.0 \right] \mps$ along the x-axis and $v_2 \in \left[-5.0, -3.0\right] \mps$ along the y-axis, to challenge approaches.
In scenario S3, a third obstacle is introduced, randomly generated from the region $x \in \left [ -3.0, -1.5\right ] \m$ and $y \in \left[-3.0, -1.5 \right] \m$ and its velocity is along the x-axis randomly generated between $v_3 \in \left[ 1.0, 3.0 \right] \mps$
The velocities of the other two obstacles are adjusted to match this range, while their directions remain unchanged.
The experimental setup is illustrated in Fig.\ref{fig:bench_setup}, and the results are summarized in Table~\ref{tab:benchmark}.

The results show that our approaches outperform others in both success rate and time-to-reach metrics. 
The observed discrepancies in path length arise because the average path length only includes successful cases, leading to simpler cases being overrepresented in SH-based methods.
The SDF-CBF-QP method fails in almost all cases due to its time-invariant safe set, which does not account for dynamic obstacles.
Furthermore, CDF-SH-QP achieves a higher success rate than SDF-SH-QP because the non-zero gradient information improves obstacle avoidance. 
Overall, our method demonstrates superior performance in highly dynamic environments in 2D scenarios.

For the 7D scenarios, we evaluate the approaches using a 7-axis Franka robot arm, as shown in Fig.~\ref{fig:7d_bench}.
In scenario C1, two obstacles are randomly generated within $x\in \left[0.3, 0.6 \right] \m$, $y \in \left[0.6, 0.9 \right] \m$ and $z\in \left[0.5, 0.7 \right] \m$, with velocities along the $y$-axis randomly selected between $-0.8 \mps$ and $-0.5 \mps$.
The initial state of the robot arm is $q(0) = (-1.09, 0.35,-0.32,-1.70, 0.18, 2.05, -0.20) \rad$ and the task is to rotate the first joint to $1.09$ without collisions.
In scenario C2, six obstacles are randomly generated within the same region, with velocities increasing across obstacles.
The first obstacle moves at a velocity range of $\left[ -0.2, -0.1\right] \mps$, the fifth at $\left[ -0.9, -1.0\right] \mps$, and the sixth covers the full range $\left[ -0.9, -0.1\right] \mps$.
The robot's task is to maintain its pose at $(0.09, 0.35,-0.32,-1.70, 0.18, 2.05, -0.20) \rad$, avoid obstacles, and return to this pose.
In scenario C3, the task remains the same as in C1, but the obstacle setup is identical to C2.

The results indicate that our method consistently outperforms others across metrics.
However, the path length is not a primary advantage, as our method prioritizes safety, resulting in slightly longer paths for simpler tasks, such as avoiding slow-moving obstacles.
In contrast, baselines tend to take more aggressive trajectories, occasionally succeeding with shorter path lengths. 
CDF-TVCBF-QP achieves a higher success rate than SDF-TVCBF-QP due to the non-zero gradient, which provides better obstacle avoidance guidance. 
While our methods account for obstacle velocities, failure can still occur due to physical limitations.

\subsection{Real-world Experiments}

To validate our approach, we conducted experiments on a 7-axis Franka robot. In our experimental setup, the right robot arm performed a goal-reaching task, moving from an initial joint configuration to a target one while accounting for a dynamic obstacle moving at a predefined speed. The left robot arm was used to simulate the moving obstacle by holding a stick with a ball, which moved at a constant velocity to ensure reproducible results. We tested obstacle velocities of 0.05 m/s and 0.15 m/s, observing distinct robot behaviors for each velocity, as shown in Fig.~\ref{fig:robot_experiments}.
Our approach demonstrates effective adaptive collision-avoidance behavior in response to moving obstacles.
When the obstacle moves slower (top), the robot efficiently reaches its target while maintaining a safe but close proximity to the moving object.
Once the obstacle moves farther away, the robot returns to the goal and stabilizes its position. Conversely, when the obstacle moves at a higher velocity (bottom), the robot adopts a more safety-focused strategy, maintaining a greater distance to account for the obstacle's faster motion and ensuring robust collision avoidance. Even with the added complexity of higher obstacle speeds, the robot successfully achieves its goal while consistently prioritizing safety.

\section{Conclusions}\label{sec:conclusion}
We presented a method for computing the velocities of moving objects in configuration space, which is subsequently used to construct TVCBFs and TVCLFs. Extensive numerical simulations, benchmarking against state-of-the-art methods, and real-world experiments demonstrate the efficacy of our approach.
The gradient information in configuration space could be further explored in future work, such as for constructing convex, collision-free sets. Additionally, we plan to investigate model predictive control-based methods and expand the scope to more elaborated whole-body manipulation planning tasks in configuration space.

{
\bibliographystyle{IEEEtran}
\bibliography{main}

\begin{thebibliography}{10}
\providecommand{\url}[1]{#1}
\csname url@samestyle\endcsname
\providecommand{\newblock}{\relax}
\providecommand{\bibinfo}[2]{#2}
\providecommand{\BIBentrySTDinterwordspacing}{\spaceskip=0pt\relax}
\providecommand{\BIBentryALTinterwordstretchfactor}{4}
\providecommand{\BIBentryALTinterwordspacing}{\spaceskip=\fontdimen2\font plus
\BIBentryALTinterwordstretchfactor\fontdimen3\font minus \fontdimen4\font\relax}
\providecommand{\BIBforeignlanguage}[2]{{%
\expandafter\ifx\csname l@#1\endcsname\relax
\typeout{** WARNING: IEEEtran.bst: No hyphenation pattern has been}%
\typeout{** loaded for the language `#1'. Using the pattern for}%
\typeout{** the default language instead.}%
\else
\language=\csname l@#1\endcsname
\fi
#2}}
\providecommand{\BIBdecl}{\relax}
\BIBdecl

\bibitem{koptev2024reactive}
M.~Koptev, N.~Figueroa, and A.~Billard, ``Reactive collision-free motion generation in joint space via dynamical systems and sampling-based mpc,'' \emph{The International Journal of Robotics Research}, p. 02783649241246557, 2024.

\bibitem{ratliff2009chomp}
N.~Ratliff, M.~Zucker, J.~A. Bagnell, and S.~Srinivasa, ``Chomp: Gradient optimization techniques for efficient motion planning,'' in \emph{2009 IEEE international conference on robotics and automation}.\hskip 1em plus 0.5em minus 0.4em\relax IEEE, 2009, pp. 489--494.

\bibitem{schulman2014motion}
J.~Schulman, Y.~Duan, J.~Ho, A.~Lee, I.~Awwal, H.~Bradlow, J.~Pan, S.~Patil, K.~Goldberg, and P.~Abbeel, ``Motion planning with sequential convex optimization and convex collision checking,'' \emph{The International Journal of Robotics Research}, vol.~33, no.~9, pp. 1251--1270, 2014.

\bibitem{spahn2021coupled}
M.~Spahn, B.~Brito, and J.~Alonso-Mora, ``Coupled mobile manipulation via trajectory optimization with free space decomposition,'' in \emph{2021 IEEE International Conference on Robotics and Automation (ICRA)}.\hskip 1em plus 0.5em minus 0.4em\relax IEEE, 2021, pp. 12\,759--12\,765.

\bibitem{williams2016aggressive}
G.~Williams, P.~Drews, B.~Goldfain, J.~M. Rehg, and E.~A. Theodorou, ``Aggressive driving with model predictive path integral control,'' in \emph{2016 IEEE International Conference on Robotics and Automation (ICRA)}.\hskip 1em plus 0.5em minus 0.4em\relax IEEE, 2016, pp. 1433--1440.

\bibitem{jankowski2023vp}
J.~Jankowski, L.~Bruderm{\"u}ller, N.~Hawes, and S.~Calinon, ``Vp-sto: Via-point-based stochastic trajectory optimization for reactive robot behavior,'' in \emph{2023 IEEE International Conference on Robotics and Automation (ICRA)}.\hskip 1em plus 0.5em minus 0.4em\relax IEEE, 2023, pp. 10\,125--10\,131.

\bibitem{koptev2021real}
M.~Koptev, N.~Figueroa, and A.~Billard, ``Real-time self-collision avoidance in joint space for humanoid robots,'' \emph{IEEE Robotics and Automation Letters}, vol.~6, no.~2, pp. 1240--1247, 2021.

\bibitem{mirrazavi2018unified}
S.~S. Mirrazavi~Salehian, N.~Figueroa, and A.~Billard, ``A unified framework for coordinated multi-arm motion planning,'' \emph{The International Journal of Robotics Research}, vol.~37, no.~10, pp. 1205--1232, 2018.

\bibitem{li2024configuration}
Y.~Li, X.~Chi, A.~Razmjoo, and S.~Calinon, ``Configuration space distance fields for manipulation planning,'' \emph{arXiv preprint arXiv:2406.01137}, 2024.

\bibitem{koptev2022neural}
M.~Koptev, N.~Figueroa, and A.~Billard, ``Neural joint space implicit signed distance functions for reactive robot manipulator control,'' \emph{IEEE Robotics and Automation Letters}, vol.~8, no.~2, pp. 480--487, 2022.

\bibitem{li2024representing}
Y.~Li, Y.~Zhang, A.~Razmjoo, and S.~Calinon, ``Representing robot geometry as distance fields: Applications to whole-body manipulation,'' in \emph{Proc. IEEE Intl Conf. on Robotics and Automation (ICRA)}, 2024, pp. 15\,351--15\,357.

\bibitem{zeng2021safety}
J.~Zeng, B.~Zhang, and K.~Sreenath, ``Safety-critical model predictive control with discrete-time control barrier function,'' in \emph{2021 American Control Conference (ACC)}.\hskip 1em plus 0.5em minus 0.4em\relax IEEE, 2021, pp. 3882--3889.

\bibitem{singletary2022safety}
A.~Singletary, W.~Guffey, T.~G. Molnar, R.~Sinnet, and A.~D. Ames, ``Safety-critical manipulation for collision-free food preparation,'' \emph{IEEE Robotics and Automation Letters}, vol.~7, no.~4, pp. 10\,954--10\,961, 2022.

\bibitem{dai2023safe}
B.~Dai, R.~Khorrambakht, P.~Krishnamurthy, V.~Goncalves, A.~Tzes, and F.~Khorrami, ``Safe navigation and obstacle avoidance using differentiable optimization based control barrier functions,'' \emph{IEEE Robotics and Automation Letters}, 2023.

\bibitem{liu2022regularized}
P.~Liu, K.~Zhang, D.~Tateo, S.~Jauhri, J.~Peters, and G.~Chalvatzaki, ``Regularized deep signed distance fields for reactive motion generation,'' in \emph{2022 IEEE/RSJ International Conference on Intelligent Robots and Systems (IROS)}.\hskip 1em plus 0.5em minus 0.4em\relax IEEE, 2022, pp. 6673--6680.

\bibitem{ames2019control}
A.~D. Ames, S.~Coogan, M.~Egerstedt, G.~Notomista, K.~Sreenath, and P.~Tabuada, ``Control barrier functions: Theory and applications,'' in \emph{2019 18th European control conference (ECC)}, 2019, pp. 3420--3431.

\bibitem{ferreau2014qpoases}
H.~J. Ferreau, C.~Kirches, A.~Potschka, H.~G. Bock, and M.~Diehl, ``qpoases: A parametric active-set algorithm for quadratic programming,'' \emph{Mathematical Programming Computation}, vol.~6, pp. 327--363, 2014.

\end{thebibliography}
}

\end{document}